\documentclass[letterpaper]{article} 
\usepackage{aaai25}  

\usepackage{times}  
\usepackage{helvet}  
\usepackage{courier}  
\usepackage[hyphens]{url}  
\usepackage{graphicx} 
\usepackage{natbib}  
\usepackage{caption} 
\usepackage{booktabs}
\usepackage{subfigure}
\usepackage{subcaption}
\usepackage{algorithm}
\usepackage{algorithmic}
\usepackage{multirow}
\usepackage{booktabs}
\usepackage{newfloat}
\usepackage{listings}
\DeclareCaptionStyle{ruled}{labelfont=normalfont,labelsep=colon,strut=off} 
\floatstyle{ruled}
\newfloat{listing}{tb}{lst}{}
\floatname{listing}{Listing}
\usepackage{amsmath}

\title{PDDLFuse: A Tool for Generating Diverse Planning Domains}
\author {
    Vedant Khandelwal,
    Amit Sheth,
    Forest Agostinelli
}
\affiliations {
    Artificial Intelligence Institute, University of South Carolina\\
    vedant@email.sc.edu, amit@sc.edu, foresta@cse.sc.edu
}

\usepackage{bibentry}

\begin{document}

\maketitle

\begin{abstract}
The variety of real-world challenges requires planning algorithms that can adapt to a broad range of domains. Traditionally, the creation of planning domains has relied heavily on human implementation, which limits the scale and diversity of available domains. While recent advancements have leveraged generative AI technologies such as large language models (LLMs) for domain creation, these efforts have predominantly focused on translating existing domains from natural language descriptions rather than generating novel ones. In contrast, the concept of domain randomization, which has been highly effective in reinforcement learning, enhances performance and generalizability by training on a diverse array of randomized new domains. Inspired by this success, our tool, PDDLFuse, aims to bridge this gap in Planning Domain Definition Language (PDDL). PDDLFuse is designed to generate new, diverse planning domains that can be used to validate new planners or test foundational planning models. We have developed methods to adjust the domain generator’s parameters to modulate the difficulty of the domains it generates. This adaptability is crucial as existing domain-independent planners often struggle with more complex problems. Initial tests indicate that PDDLFuse efficiently creates intricate and varied domains, representing a significant advancement over traditional domain generation methods and making a contribution towards planning research.
\end{abstract}

%

\section{Introduction}

Automated planning systems are critical in applications ranging from robotics to software management. These systems depend on well-defined planning domains and problems that describe the  environment and define the tasks \citep{ghallab2004automated}. Traditionally, planning domains have been manually created, which restricts the diversity and complexity of the problems these systems can tackle. This limitation affects the robustness and generalizability of planning algorithms when they encounter new or unforeseen domains \citep{chen2024learning}.

Recent advancements in generative AI, particularly large language models (LLMs), have been used for automating the creation of planning domains. While these technologies have the potential to scale domain creation beyond traditional methods, most applications of LLMs in this area have been limited to translating existing domains from natural language descriptions rather than generating novel and varied domains \cite{oswald2024large}. This replication restricts the ability of planning systems to generalize effectively across a broad set of domains.

Domain randomization, is proven to enhance performance and generalization in reinforcement learning, involves training agents in varied domains. Successfully applied in fields such as robotic control and locomotion, this method demonstrates that training in diverse settings can significantly improve an agent’s ability to adapt to new, unseen domains \citep{mehta2020active, ajani2023evaluating}. Inspired by the success of domain randomization in reinforcement learning, which has shown improved adaptability and robustness, we propose applying similar principles within the Planning Domain Definition Language (PDDL).

PDDLFuse is a tool designed to generate new planning domains by fusing existing ones, rather than merely creating existing domains through translation from natural language descriptions \citep{oswald2024large, mahdavi2024leveraging}. This approach enhances the diversity of domains available for planning research and supports the development of more adaptable and generalizable planning algorithms. By significantly expanding the range of test domains, PDDLFuse facilitates comprehensive testing and development of planning algorithms, validation of new planners, testing foundational planning models, and exploring previously uncharted domains. Preliminary tests indicate that PDDLFuse efficiently generates complex and diverse domains, representing an advancement over traditional methods and contributing to the field of planning research.

The following sections will discuss the background of planning domains, and review related works, describe the methodologies employed, and present experimental results.

\section{Background and Related Works}

This section covers the essentials of planning domains, domain-independent planners, and the role of generative AI in domain reconstruction. We highlight the limitations of current methods. We also explore domain randomization, an approach from reinforcement learning that improves algorithm robustness by training with diverse domains, offering potential benefits for planning systems. Detailed discussion in \textit{Supplementary Material}.

\subsection{Planning Domains and Problems}

In the context of automated planning, a \textit{planning domain} is a structured description of an environment consisting of objects, predicates, and actions that an agent can perform. Formally, a planning domain $\mathcal{D}$ is defined by a tuple $(\mathcal{O}, \mathcal{P}, \mathcal{A})$, where:
\begin{itemize}
    \item $\mathcal{O}$ is a finite set of \textit{objects} that exist within the domain.
    \item $\mathcal{P}$ is a finite set of \textit{predicates}, where each predicate represents a property or relationship among objects (e.g., \texttt{At(location, package)}).
    \item $\mathcal{A}$ is a finite set of \textit{actions}, where each action $a \in \mathcal{A}$ is defined by a pair $(\text{pre}(a), \text{eff}(a))$:
    \begin{itemize}
        \item $\text{pre}(a)$, the \textit{preconditions} of action $a$, is a set of predicates that must hold true for $a$ to be executed.
        \item $\text{eff}(a)$, the \textit{effects} of action $a$, is a set of predicates that describe the changes in the domain after $a$ is executed.
    \end{itemize}
\end{itemize}

A \textit{planning problem} specifies a particular task within a domain by defining both an initial and a goal state. Formally, a planning problem $\mathcal{P}$ is defined by the tuple $(\mathcal{D}, s_0, s_g)$, where:
\begin{itemize}
    \item $\mathcal{D}$ is the domain in which the problem is defined.
    \item $s_0$ is the \textit{initial state}, a set of grounded predicates representing the domain file’s state before planning begins.
    \item $s_g$ is the \textit{goal state}, a set of grounded predicates specifying the desired conditions that define the successful completion of the task.
\end{itemize}

\subsection{Domain Reconstruction}

Automating domain creation through translation from natural language has seen progress, though it still faces limitations in fostering diversity and novelty. \citet{oswald2024large} employed LLMs to replicate planning domains from textual descriptions, closely aligning with existing PDDL specifications, but requiring a reference domain for validation restricts its scope to known domains. Similarly, \citet{mahdavi2024leveraging} used an iterative refinement approach with environment feedback to enhance LLM-generated domains, reducing manual effort but focusing primarily on refining rather than creating new domains, limiting scalability. The “Translate-Infer-Compile” (TIC) tool by \citet{agarwal2024tic} and “AUTOPLANBENCH” by \citet{steinaautomating} advance the translation of natural language into structured PDDL, enhancing accuracy through logic reasoning and LLM interaction, yet they remain confined to reconstructing established domains, rather than diversifying the domain pool essential for generalization in planning.

\subsection{Generalization in Planning}

Generalization in automated planning is limited by the lack of diverse training domains, as exemplified by those in the International Planning Competition (IPC), leading to weaker inductive biases and models prone to overfitting. Recent approaches using Graph Neural Networks (GNNs) and Large Language Models (LLMs) aim to address this but remain restricted by their narrow domain scope. For instance, the graph representations by \citet{chen2024learning} and the GOOSE tool by \citet{chen2023goose} show promise in learning domain-independent heuristics, yet face scalability issues and reliance on IPC domains. Similarly, \citet{toyer2018action} employs GNNs to improve plan quality but depends on accessible solvers, limiting real-world applicability. LLMs, even with advanced prompting techniques \citep{hu2023chain, yao2023tree}, are constrained by a lack of domain diversity, reducing their effectiveness in novel contexts. Multimodal and code-based models, such as those by \citet{lu2023multimodal, pallagani2023plansformer, khandelwal2024towards, agostinelli2024specifying}, perform well in known distributions but struggle with out-of-distribution domains, highlighting the need for more diverse domains to achieve robust generalization.

\subsection{Domain Randomization and Generalization in Reinforcement Learning}

Domain randomization is a key technique in reinforcement learning (RL) that is used to improve robustness and generalizability by exposing agents to diverse training domains, thereby improving generalization to unfamiliar domains. \citet{mehta2020active} introduced Active Domain Randomization (ADR), which strategically manipulates challenging environmental parameters to improve policy robustness, particularly in robotic control. Similarly, \citet{ajani2023evaluating} showed that varying physical properties like surface friction can significantly boost an RL agent’s generalization ability. \citet{kang2024balanced} further refined this with Balanced Domain Randomization (BDR), which focuses training on rare and complex domains to enhance performance under demanding conditions. In non-physical tasks, \citet{koo2019adversarial} applied adversarial domain adaptation to align feature representations across domains, thereby enhancing policy generalization in complex tasks like dialogue systems. These studies demonstrate the power of domain randomization in building resilient AI, aligning with PDDLFuse’s goal to generate diverse planning domains to improve generalization in automated planning.

\subsection{Domain Independent Planners}
\label{subsec:domain-independent-planners-bg}

Domain-independent planners such as Fast Downward (FD)\citep{helmert2006fast} and LPG \citep{gerevini2002lpg} play a crucial role in the advancement of automated planning technologies. These systems are designed to function across a wide range of problem domains by utilizing heuristics that do not rely on specific domain knowledge. FD converts PDDL tasks into a more manageable internal format and employs powerful heuristics like the Fast-Forward (FF) \citep{hoffmann2001ff}, which simplifies planning by focusing only on positive action effects, and the landmark-cut (lmcut) \citep{helmert2009landmarks}, which identifies essential milestones within a plan to optimize the search process. LPG, on the other hand, leverages stochastic local search strategies that incrementally refine plans through action-graph and plan-graph techniques, proving highly effective in both propositional and numerical planning contexts \citep{gerevini2002lpg}. The use of these planners in research is driven by their ability to efficiently generate solutions in diverse domains, thereby facilitating the development of more robust and adaptable planning systems.

\section{Methods}

This section details the procedures and algorithms developed to fuse existing domains and manipulate domain characteristics to generate new and diverse domains. More details in \textit{Supplementary Material.}

\subsection{Domain Generation}
In PDDLFuse, the generation of new planning domains $\mathcal{D} = (\mathcal{O}, \mathcal{P}, \mathcal{A})$ starts by selecting two existing domains and their problem files as bases. An initial step ensures no overlapping predicates or action names exist between the two domains, by systematically renaming the predicates and actions to maintain uniqueness.

The actions within the domains are then enhanced using a set of hyperparameters that control modifications to preconditions and effects. These parameters include:
\begin{itemize}
    \item Probability of adding a new predicate to the preconditions ($\text{prob\_add\_pre}$).
    \item Probability of adding a new effect to the actions ($\text{prob\_add\_eff}$).
    \item Probability of removing a predicate from the preconditions ($\text{prob\_rem\_pre}$).
    \item Probability of removing a predicate from the effects ($\text{prob\_rem\_eff}$).
\end{itemize}
The inclusion of negations through $\text{prob\_neg}$, ensuring predicate reversibility $rev_flag$ and the control over object counts via $\text{num\_objs}$ provide further flexibility. This process allows for the generation of complex and diverse planning domains, essential for developing robust planning algorithms.

\subsection{Problem File Generation}
Problem file generation starts with setting the initial state based on $\text{num\_objs}$. A sequence of random actions from the generated domain is executed to transition the initial state into a new state, where a subset of true predicates forms the goal state. This process ensures the generated problems are solvable within the domain.

\begin{algorithm}
\small
\caption{Domain \& Problem Generation}
\begin{algorithmic}[1]
\REQUIRE Two base domains $\mathcal{D}_1 = (\mathcal{O}_1, \mathcal{P}_1, \mathcal{A}_1)$ and $\mathcal{D}_2 = (\mathcal{O}_2, \mathcal{P}_2, \mathcal{A}_2)$
\ENSURE No overlapping predicates or actions between $\mathcal{D}_1$ and $\mathcal{D}_2$
\STATE $\mathcal{O} \gets \mathcal{O}_1 \cup \mathcal{O}_2$ \COMMENT{Union of objects from both domains}
\STATE $\mathcal{P} \gets \mathcal{P}_1 \cup \mathcal{P}_2$ \COMMENT{Union of predicates from both domains}
\STATE $\mathcal{A} \gets \emptyset$
\FOR{each action $a$ in $\mathcal{A}_1 \cup \mathcal{A}_2$}
    \STATE Define $\text{pre}(a)$ and $\text{eff}(a)$ for new $\mathcal{A}$
    \IF{random() $< \text{prob\_add\_pre}$}
        \STATE Add new predicates to $\text{pre}(a)$ \COMMENT{Expanding precond}
    \ENDIF
    \IF{random() $< \text{prob\_add\_eff}$}
        \STATE Add new effects to $\text{eff}(a)$ \COMMENT{Expanding effects}
    \ENDIF
    \IF{random() $< \text{prob\_rem\_pre}$}
        \STATE Remove predicates from $\text{pre}(a)$ \COMMENT{Simplifying precond}
    \ENDIF
    \IF{random() $< \text{prob\_rem\_eff}$}
        \STATE Remove predicates from $\text{eff}(a)$ \COMMENT{Simplifying effects}
    \ENDIF
    \STATE Apply $\text{prob\_neg}$ to negate added predicates
    \STATE $\mathcal{A} \gets \mathcal{A} \cup \{a\}$ \COMMENT{Incorporating modified action into new domain}
\ENDFOR
\STATE Generate problem $\mathcal{P}$ using modified $\mathcal{D}_{new} = (\mathcal{O}, \mathcal{P}, \mathcal{A})$ and $\text{num\_objs}$
\STATE Execute actions to derive the goal state from the initial state
\RETURN $\mathcal{D}_{new}$, $\mathcal{P}$ \COMMENT{Output new planning domain and problem}
\end{algorithmic} \label{alg:algorithm}
\end{algorithm}

 This section outlined the systematic approach employed by PDDLFuse to generate new and diverse planning domains, with an emphasis on parametric controls that enables customization. The subsequent sections will discuss the experimental setup for evaluating the effectiveness and utility of these generated domains in planning research.

\section{Results}
This section presents the outcomes of experiments conducted to evaluate PDDLFuse's ability to generate novel planning domains and assess the solvability of these domains by domain-independent planners. We explore the performance of FD and LPG planners across various domain generation parameters, providing insights into their limitations and the complexity of the generated domains. All planners ran with a time limit of 200 seconds per problem instance.

\subsection{Planners and Heuristics Evaluation}

In our planning experiments, we utilized two domain-independent planners: Fast Downward (FD) with FF and lmcut heuristics, and LPG.
\paragraph{Experimental Setup and Parameters}
For these experiments, we introduced variation within the generated domains using the following parameters:
\begin{itemize}
    \item \texttt{Prob\_add\_precond} = 0.5
    \item \texttt{Prob\_add\_effect} = 0.5
    \item \texttt{Prob\_remove\_effect} = 0.3
    \item \texttt{Prob\_negation\_predicate} = 0.5
\end{itemize}

\begin{figure*}[!htbp]
    \centering
    \includegraphics[width=0.85\textwidth]{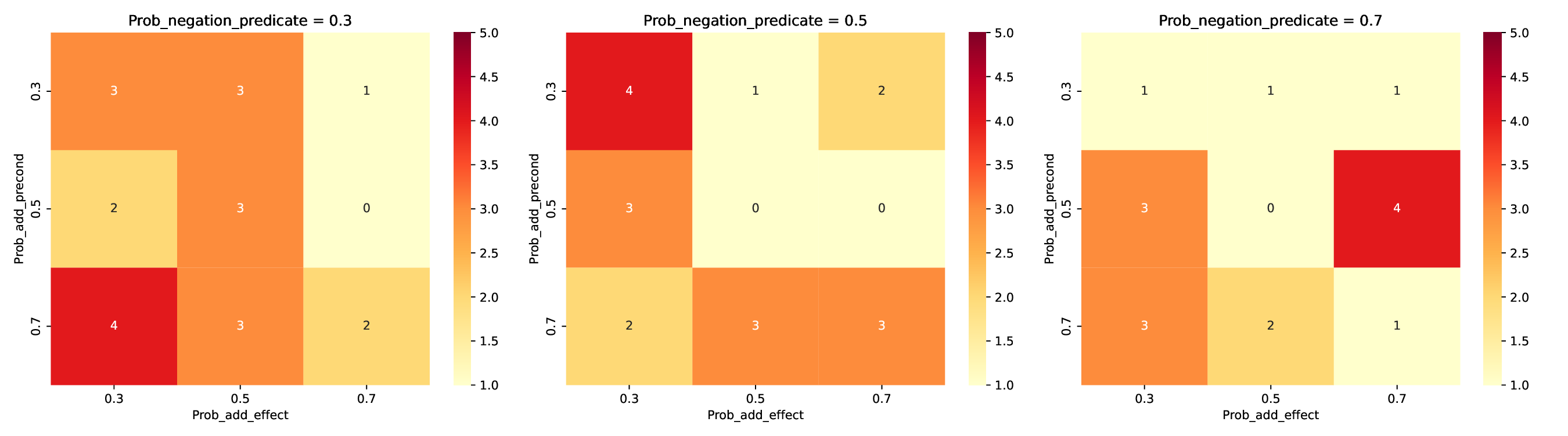}
    \caption{Solvability heat maps for domains with 15 objects, evaluated using the FD(FF) across varied parameter configurations.}
    \label{fig:heat_maps}
\end{figure*}

Using these parameters, we generated ten domains per depth level. Starting from Gripper and Blocks World as base for depth level 1, subsequent levels used the previous depth’s generated domains as base. This setup systematically assessed planner performance across increasing domain complexities. These generated domains are used to evaluate planner solvability across increasing domain depths. We measured the planners’ success rates, and path costs. These metrics provided a comprehensive view of planner efficiency and adaptability as domain complexity increased.

\begin{table}[h]
\centering
\caption{Planner Performance Across Depth Levels}
\renewcommand{\arraystretch}{0.8} 
\begin{tabular}{p{1cm} p{1.5cm} p{1.5cm} p{1cm} p{1cm}}
\toprule
\textbf{Depth} & \textbf{Planner} & \textbf{Solved} & \textbf{Path Cost}  \\
\midrule
\multirow{2}{*}{1} & LPG  & 10/10 & 15 \\
                   & FD(FF) & 10/10 & 15 \\
\midrule
\multirow{2}{*}{2} & LPG  & 8/10  & 11\\
                   & FD(FF) & 10/10 &  12\\
\midrule
\multirow{2}{*}{3} & LPG  & 7/10  &  8\\
                   & FD(FF) & 9/10  &  10\\
\midrule
\multirow{2}{*}{4} & LPG  & 5/10  &  8\\
                   & FD(FF) & 8/10  &  9\\
\midrule
\multirow{2}{*}{5} & LPG  & 4/10  &  7\\
                   & FD(FF) & 8/10  &  9\\
\bottomrule
\end{tabular}
\label{tab:depth_sol}
\end{table}

As shown in Table \ref{tab:depth_sol}, both FD(FF) and LPG maintain high solvability at lower depth levels, but their performance declines as depth increases. LPG shows robustness across depths, though its effectiveness drops as depth increases, while FD(FF) maintains consistent performance in more complex domains. The recorded path costs further demonstrate the computational impact of domain depth, with values rising as domain complexity increases.

\subsection{Solvability Across Parameter Variations}

We used five base domains—Blocks-World, Gripper, Depot, Grid, and Satellite—to conduct experiments focusing on depth level 1 with 15 objects. Other parameters were systematically varied to examine planner performance under different domain configurations. The parameter pairs used were:
\begin{itemize}
    \item \texttt{Prob\_add\_precond} and \texttt{Prob\_remove\_precond}: (0.3, 0.7), (0.5, 0.5), (0.7, 0.3)
    \item \texttt{Prob\_add\_effect} and \texttt{Prob\_remove\_effect}: (0.3, 0.7), (0.5, 0.5), (0.7, 0.3)
\end{itemize}

Additionally, we varied the negation probability with values of 0.3, 0.5, and 0.7, enabling predicate reversibility throughout the experiments.

\paragraph{Heat Map Analysis}
The heat maps in Figure \ref{fig:heat_maps} show FD(FF) success rates across different parameter configurations. Each cell’s value represents planner solvability for specific combinations of add/remove probabilities and negation values. Results indicate that moderate negation probabilities and balanced add/remove probabilities generally yield higher solvability, whereas extreme configurations (e.g., high add/remove probabilities or high negation) decrease success rates. These findings underscore the planner's sensitivity to parameter variations, revealing configurations that increase complexity and pose challenges for domain-independent planning. Additional results are available in the \textit{Supplementary Material.}

\subsection{System Configuraiton}
Experiments were conducted on CPU-only nodes, equipped with Intel Xeon Platinum 8480CL processors, featuring two sockets with 56 cores each, totaling 112 cores per node. Each node was allocated 500GB of RAM. For each experiment, 14 cores were utilized, providing approximately 608MB of RAM per experiment with a random seed of 42.



\section{Conclusion and Future Work}
In this work, we presented PDDLFuse, a tool designed to generate diverse and complex planning domains that can be used to train more robust planning foundation models. The generated domains serve as a valuable resource for testing and benchmarking new planning algorithms, pushing the boundaries of current domain-independent planners like Fast Downward and LPG-td. Our experiments demonstrated the tool’s ability to generate a wide variety of domains, with results showing that increased domain complexity significantly challenges existing planning systems. This analysis highlights the impact of varied parameter settings on planner performance, underscoring the potential of PDDLFuse to generate complex domains.

Future work involves integrating feedback mechanisms to adjust parameters based on planner performance dynamically, which could further refine domain complexity, advance the development of more resilient and adaptable AI planning systems, and further develop a comprehensive, adaptive tool that generates planning domains by adjusting a predefined set of parameters, complexity levels, and target characteristics. This enhanced system would create diverse domains and replicate existing ones by precisely tuning parameters, thereby broadening its applicability. Additionally, it would provide interpretative insights into each domain’s structure, improving usability. 

\bibliography{aaai25}

\newpage
\appendix
\section{Background and Related Works}
\label{appendix:related_works}
In this section, we explore foundational concepts and review significant advancements related to planning domains, generalization in planning, and domain generation techniques. We discuss traditional approaches to domain creation, the use of large language models for domain reconstruction, and challenges related to generalization in planning tasks. Additionally, we highlight domain randomization techniques in reinforcement learning and examine key domain-independent planners, such as Fast Downward and LPG, including the heuristics that support their functionality. This background provides a comprehensive understanding of the context and motivations behind developing our PDDLFuse tool for generating diverse and complex planning domains. 

\subsection{Planning Domains and Problems}

In the context of automated planning, a \textit{planning domain} is a structured description of an environment consisting of objects, predicates, and actions that an agent can perform. Formally, a planning domain $\mathcal{D}$ is defined by a tuple $(\mathcal{O}, \mathcal{P}, \mathcal{A})$, where:
\begin{itemize}
    \item $\mathcal{O}$ is a finite set of \textit{objects} that exist within the domain.
    \item $\mathcal{P}$ is a finite set of \textit{predicates}, where each predicate represents a property or relationship among objects (e.g., \texttt{At(location, package)}).
    \item $\mathcal{A}$ is a finite set of \textit{actions}, where each action $a \in \mathcal{A}$ is defined by a pair $(\text{pre}(a), \text{eff}(a))$:
    \begin{itemize}
        \item $\text{pre}(a)$, the \textit{preconditions} of action $a$, is a set of predicates that must hold true for $a$ to be executed.
        \item $\text{eff}(a)$, the \textit{effects} of action $a$, is a set of predicates that describe the changes in the environment after $a$ is executed.
    \end{itemize}
\end{itemize}

A \textit{planning problem} specifies a particular task within a domain by defining both an initial and a goal state. Formally, a planning problem $\mathcal{P}$ is defined by the tuple $(\mathcal{D}, s_0, s_g)$, where:
\begin{itemize}
    \item $\mathcal{D}$ is the domain in which the problem is defined.
    \item $s_0$ is the \textit{initial state}, a set of grounded predicates representing the environment’s state before planning begins.
    \item $s_g$ is the \textit{goal state}, a set of grounded predicates specifying the desired conditions that define the successful completion of the task.
\end{itemize}

Given a domain $\mathcal{D}$ and problem $\mathcal{P}$, a \textit{plan} $\pi$ is a sequence of actions $(a_1, a_2, \dots, a_n)$ that, when applied in order from $s_0$, transitions the initial state to a state where $s_g$ holds true. The objective of a planner is to find a valid plan $\pi$ that satisfies $s_g$, optimizing for factors such as plan length or action cost in some cases.

The structured nature of planning domains and problems allows automated planners to systematically search for solutions, making planning a core component of decision-making systems in AI.


\subsection{Domain Reconstruction}
Several studies have focused on automating domain generation by reconstructing existing domains from natural language descriptions. For example, \citet{oswald2024large} proposed a framework where LLMs are used to reconstruct planning domains from textual descriptions, aligning closely with ground-truth PDDL specifications. However, this approach requires a reference domain for validation, restricting its applicability to pre-existing domains and limiting its capacity to foster novel, unencountered domains. This reliance on predefined standards constrains the generation of diverse domains necessary for improved generalization in planning.

Similarly, \citet{mahdavi2024leveraging} introduced an iterative refinement process that leverages environment feedback to enhance LLM-generated PDDL domains. While this approach reduces human intervention, it remains heavily reliant on environmental validation and is focused on modifying existing structures rather than generating new domains. The need for accessible validation functions also limits its scalability.

The “Translate-Infer-Compile” (TIC) framework by \citet{agarwal2024tic} translates natural language descriptions into structured intermediate representations, later compiled into PDDL task files using a logic reasoner. This method improves translation accuracy but is limited to domain reconstruction, making it unsuitable for creating diverse, novel domains. \citet{steinaautomating} also proposed “AUTOPLANBENCH” to convert PDDL files into natural language prompts for LLM-based action choice, enhancing LLM interaction but without expanding the pool of available domains beyond pre-existing templates.

While these studies contribute valuable methods for domain reconstruction, they cannot generate new, diverse domains essential for robust planning generalization. DomGenX addresses this gap by generating novel domains independent of existing templates, significantly enhancing the diversity of available domains. This expanded domain pool enables the training of planning models with stronger inductive biases, enhancing adaptability across a wider range of real-world challenges.

\subsection{Generalization in Planning}

Achieving generalization in planning remains challenging, largely due to the limited diversity in available training domains, such as those in the International Planning Competition (IPC). This lack of domain variety results in weaker inductive biases within machine learning models, as they learn patterns from a narrow training set rather than developing broadly applicable knowledge. Robust inductive bias, which enables models to recognize generalizable patterns across diverse domains, is essential for effective adaptation to new and varied domains. Without this, models are prone to overfitting, limiting their adaptability in real-world applications.

Recent works have attempted to improve generalization by using Graph Neural Networks (GNNs) and Large Language Models (LLMs), though these approaches remain constrained by the limited set of domains. For instance, \citet{chen2024learning} proposed novel graph representations to learn domain-independent heuristics using GNNs. While effective, this approach faces scalability challenges with large graphs and relies on grounded representations that limit flexibility. Similarly, the GOOSE framework by \citet{chen2023goose} uses GNNs for learning heuristics, but its reliance on IPC domains restricts its generalization potential.

\citet{toyer2018action} explored GNN-based heuristics aimed at improving plan quality, though their method’s dependence on supervised learning with moderately challenging instances assumes accessible solvers, which is impractical for many real-world domains. Likewise, LLMs pre-trained on large-scale text datasets have shown limited success in planning contexts. Despite advancements in prompting techniques \cite{hu2023chain, yao2023tree}, LLMs still lack the domain diversity needed for robust generalization.

Studies on multimodal and code-based models, such as those by \citet{lu2023multimodal} and \citet{pallagani2023plansformer}, similarly reveal that these models, though effective on in-distribution tasks, struggle to generalize to out-of-distribution domains, further underscoring the need for broader domain diversity to support generalization across varied settings.

Our work introduces DomGenX, a novel domain and problem generator, to address this gap. Unlike prior methods restricted to finite sets of domains, DomGenX combines existing domain files to generate a broad range of novel, randomized domains, substantially increasing the diversity of training data. By training on this expanded set of domains, DomGenX fosters stronger inductive biases, enhancing the potential for planning systems to generalize effectively across diverse and unseen problem domains. This approach ultimately provides a foundation for more adaptable and robust planning algorithms.

\subsection{Domain Randomization and Generalization in Reinforcement Learning}

Domain randomization has proven to be a powerful tool in reinforcement learning (RL). It trains agents across diverse environments, enhancing their adaptability and robustness.

\citet{mehta2020active} introduced Active Domain Randomization (ADR) to tackle high-variance policies seen in zero-shot transfer. By selecting challenging environment parameters, ADR focuses training on difficult domains, improving policy robustness in tasks such as robotic control. In locomotion tasks, \citet{ajani2023evaluating} evaluated domain randomization by varying environmental parameters like surface friction, showing that specific randomization improves RL agents’ generalization to unseen environments, reinforcing the importance of controlled diversity for real-world transfer. \citet{kang2024balanced} proposed Balanced Domain Randomization (BDR) to address training imbalances by emphasizing rare, challenging domains. This method enhances worst-case performance, making agents more robust in unpredictable settings, which underscores the value of diverse training conditions. Lastly, \citet{koo2019adversarial} used adversarial domain adaptation for RL in dialog systems to align feature representations across domains, thus improving policy generalization in complex, non-physical tasks.

These studies collectively illustrate that training over diverse domains strengthens generalization across RL tasks by diversifying training contexts, aligning with DomGenX’s goal to generate domains for enhanced generalization in planning.

\subsection{Domain Independent Planner}
\subsubsection{Fast Downward Planner.}

Fast Downward is a highly versatile planning system that operates based on the planning graph concept, employing a more efficient representation known as multi-valued planning tasks. Developed by Torsten Helmert, Fast Downward has been prominent in the planning community and successful in several planning competitions. It efficiently translates PDDL (Planning Domain Definition Language) tasks into a compact internal representation, facilitating more effective planning solutions. The system's modularity allows for the application of various search algorithms and heuristics, tailored to specific problem types.

\begin{itemize}
    \item \textbf{FF Heuristic }:
    The FF heuristic, or "Fast-Forward" heuristic, is central to the Fast Downward planner, developed by Joerg Hoffmann and Bernhard Nebel. It is recognized for its rapid and effective planning capabilities, mainly due to its approach of ignoring the delete lists of actions, which simplifies the search process significantly. This heuristic generates relaxed plans by considering only the positive effects of actions, enabling quick heuristic calculations and efficient plan generation.
\item \textbf{lmcut Heuristic }:
The landmark-cut (lmcut) heuristic, another innovative heuristic used within the Fast Downward framework, calculates the minimum cost of achieving all necessary landmarks in a planning task. A landmark is a fact or a set of facts that must be true at some point in every valid plan. This heuristic identifies critical paths and bottlenecks in the plan's causal graph, aiding in the formulation of more efficient solutions.
\end{itemize}

\subsubsection{LPG Planner.}

The LPG (Local Search for Planning Graphs) planner utilizes stochastic local search techniques to effectively handle propositional and numerical planning problems. Developed by Alfonso Gerevini and Ivan Serina, LPG iteratively refines a candidate plan through a combination of action-graph refinement and plan-graph expansion, showcasing robust performance across diverse domains.

\section{Methods}
In this section, we provide detailed descriptions of the steps used within PDDLFuse to ensure the generation of diverse and solvable planning domains. The methods outlined include handling overlapping predicate and action names when combining two domains, dynamically generating action sequences to simulate goal states, and validating the generated domains and problems against PDDL 3.1 standards. These processes enhance the robustness and generalizability of generated domains, providing a reliable foundation for testing planning algorithms. Detailed algorithms for each method are discussed below.

\subsection{Handling Overlapping Predicate and Action Names}
\label{appendix:overlap_handling}

Handling overlapping predicates and action names is crucial for maintaining the integrity and uniqueness of domain definitions when combining two existing domains. Overlapping elements can cause logical conflicts and inaccuracies in domain behavior during planning tasks. This algorithm identifies overlaps and systematically renames the conflicting elements in one domain to prevent ambiguity.

\begin{algorithm}
\caption{Handling Overlapping Names}
\begin{algorithmic}[1]
\REQUIRE Two domains $\mathcal{D}_1$ and $\mathcal{D}_2$ with potential overlapping predicates and actions.
\ENSURE Unique predicates and actions across $\mathcal{D}_1$ and $\mathcal{D}_2$.
\STATE Initialize set $\mathcal{O}_1$ for unique objects from $\mathcal{D}_1$.
\STATE Initialize set $\mathcal{O}_2$ for unique objects from $\mathcal{D}_2$.
\FOR{each predicate or action in $\mathcal{D}_1$ and $\mathcal{D}_2$}
    \IF{exists in both $\mathcal{D}_1$ and $\mathcal{D}_2$}
        \STATE Rename in $\mathcal{D}_2$.
    \ENDIF
\ENDFOR
\RETURN Updated $\mathcal{D}_1$ and $\mathcal{D}_2$ with unique names.
\end{algorithmic}
\end{algorithm}

\appendix

\subsection{Action Generator Process}
\label{appendix:action_generation}

The Action Generator is designed to dynamically produce a sequence of actions based on the specifications of a given domain's initial state. It simulates random actions from the domain's action set, transforming the initial state into a new state that can potentially serve as a goal state for planning problems. The goal state is determined by selecting a subset of predicates active in the reached state, ensuring that there exists a sequence of actions that leads to this state.

\begin{algorithm}
\caption{Action Generator Process}
\begin{algorithmic}[1]
\REQUIRE Generated domain $\mathcal{D} = (\mathcal{O}, \mathcal{P}, \mathcal{A})$, Initial state $s_0$
\STATE Initialize $\text{current\_state} \gets s_0$
\STATE Initialize $\text{action\_sequence} \gets []$
\FOR{$i = 1$ to $N$}
    \STATE Select $a \in \mathcal{A}$ randomly such that preconditions of $a$ are satisfied in $\text{current\_state}$
    \STATE Apply $a$ to $\text{current\_state}$
    \STATE Append $a$ to $\text{action\_sequence}$
    \STATE Update $\text{current\_state}$ based on the effects of $a$
\ENDFOR
\STATE Define goal state $s_g$ as a subset of predicates true in $\text{current\_state}$
\RETURN $\text{action\_sequence}$, $s_g$
\end{algorithmic}
\end{algorithm}

\subsection{Validator Process}
\label{appendix:validator_process}

The Validator ensures that the generated domains and problems conform to PDDL 3.1 standards, utilizing a parser to validate the structural and logical correctness of the domain and problem definitions. It checks for consistency in the domain's actions and the feasibility of achieving the problem's goal state from its initial state based on the defined actions.

\begin{algorithm}
\caption{Validation Process}
\begin{algorithmic}[1]
\REQUIRE Domain $\mathcal{D} = (\mathcal{O}, \mathcal{P}, \mathcal{A})$, Problem $\mathcal{P} = (\mathcal{D}, s_0, s_g)$
\STATE Parse $\mathcal{D}$ and $\mathcal{P}$ using a PDDL 3.1 parser
\STATE Check for syntactical correctness of $\mathcal{D}$ and $\mathcal{P}$
\STATE Verify logical consistency: all actions in $\mathcal{A}$ must correctly transform predicates from $s_0$ to achieve $s_g$
\IF{all checks pass}
    \STATE Return "Validation Successful"
\ELSE
    \STATE Return "Validation Failed"
\ENDIF
\end{algorithmic}
\end{algorithm}

\section{Results}
In this section, we present the outcomes of our experiments, evaluating the PDDLFuse tool’s efficacy in generating complex planning domains and the performance of domain-independent planners across various configurations. We assess the robustness of our validator, analyze planner solvability under different parameter variations, and examine planner performance across increasing domain depths. These results underscore the versatility and challenge of the generated domains, demonstrating the value of PDDLFuse for advancing research in automated planning. Detailed findings are discussed below.

\subsection{Evaluating the Validator}

To assess the robustness and accuracy of our validator, we conducted evaluations across eight diverse domains, each with 20 problem instances. The problem data was sourced from the study by \citet{chen2023goose}, which used the Scorpion planner to generate optimal plans for each domain and problem file. This evaluation focused on two primary aspects: (1) verifying that the validator could correctly interpret and check the syntax of both the domain and problem files, and (2) ensuring that it could execute each optimal plan step-by-step to reach the specified goal state.

\begin{table}[ht]
    \centering
    \caption{Validator Performance Across Domains}
    \label{tab:validator_results}
    \begin{tabular}{|l|c|c|}
        \hline
        \textbf{Domain} & \textbf{Reached Goal State} & \textbf{Success Rate}\\
        \hline
        Blocks    & Yes & 100\%\\
        Ferry     & Yes & 100\% \\
        Gripper   & Yes & 100\%\\
        N-Puzzle  & Yes & 100\%\\
        Sokoban   & Yes & 100\%\\
        Spanner   & Yes & 100\%\\
        Visitall  & Yes & 100\%\\
        Visitsome & Yes & 100\%\\
        \hline
    \end{tabular}
\end{table}

As summarized in Table \ref{tab:validator_results}, the validator demonstrated perfect performance, successfully reaching the goal state in 100\% of cases across all tested domains. Each optimal plan was validated without error, affirming the validator's ability to reliably execute complex action sequences across a range of domain types.

This experiment underscores the validator’s essential role in confirming both the syntactic integrity and operational feasibility of plans within diverse domains. By accurately verifying optimal plans across varied problem sets, the validator supports robust evaluation and validation of planning solutions, ensuring that generated plans align with intended outcomes across different planning contexts. This high level of reliability is critical for advancing automated planning systems that depend on accurate and interpretable validation processes.

\subsection{Solvability Across Parameter Variations}

We used five base domains—Blocks-World, Gripper, Depot, Grid, and Satellite—to conduct experiments focusing on depth level 1 with 15 objects. Other parameters were systematically varied to examine planner performance under different domain configurations. The parameter pairs used were:
\begin{itemize}
    \item \texttt{Prob\_add\_precond} and \texttt{Prob\_remove\_precond}: (0.3, 0.7), (0.5, 0.5), (0.7, 0.3)
    \item \texttt{Prob\_add\_effect} and \texttt{Prob\_remove\_effect}: (0.3, 0.7), (0.5, 0.5), (0.7, 0.3)
\end{itemize}

Additionally, we varied the negation probability with values of 0.3, 0.5, and 0.7, enabling predicate reversibility throughout the experiments.

\begin{figure*}[!t]
    \centering
    \includegraphics[width=\textwidth]{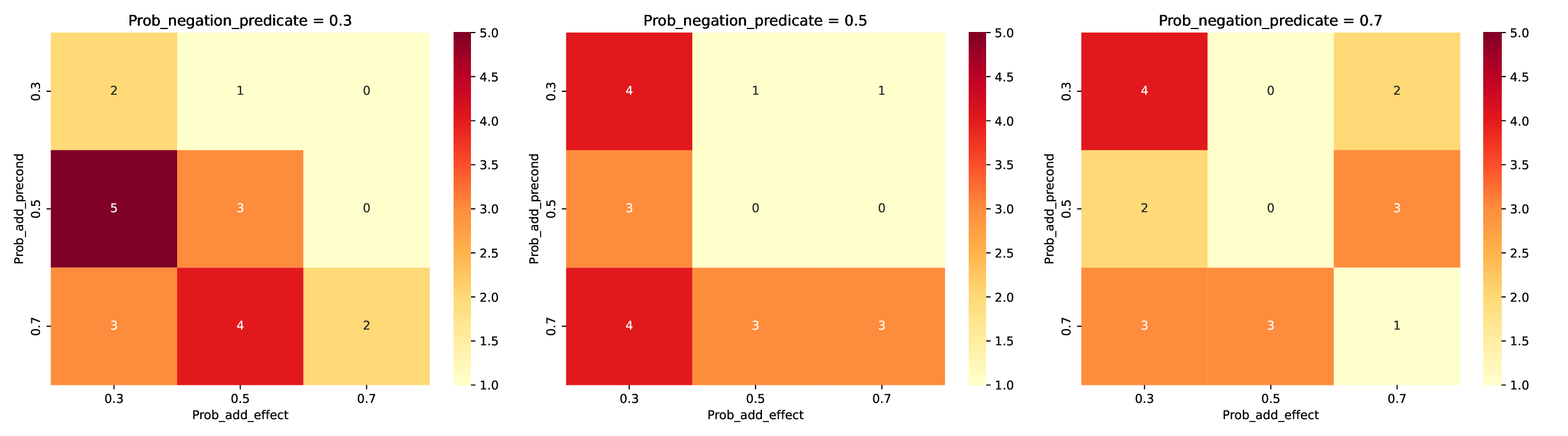}
    \caption{Solvability heat maps for domains with 15 objects, evaluated using the Fast Downward planner with lmcut heuristic across varied parameter configurations.}
    \label{fig:heat_maps_lmcut}
\end{figure*}

\begin{figure*}[!t]
    \centering
    \includegraphics[width=\textwidth]{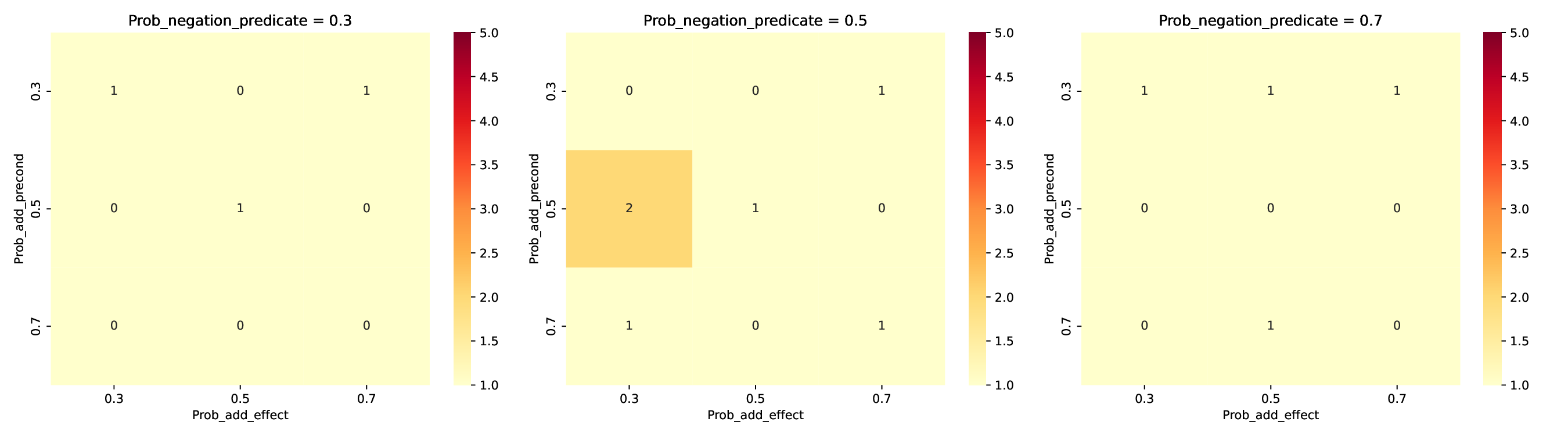}
    \caption{Solvability heat maps for domains with 15 objects, evaluated using the LPG Planner across varied parameter configurations.}
    \label{fig:heat_maps_lpg}
\end{figure*}

\paragraph{Figure \ref{fig:heat_maps_lmcut} Analysis:} The solvability heat maps illustrate the performance of the Fast Downward planner with the lmcut heuristic on domains with 15 objects across varied parameter configurations. Each cell’s value indicates the number of solvable instances within a specific configuration, combining different probabilities for adding and removing preconditions and effects, as well as varying negation probabilities. Results indicate that configurations with balanced add/remove probabilities and moderate negation values yield better solvability, while more extreme parameter settings pose greater challenges. This highlights the limitations of the lmcut heuristic in handling high-complexity domains generated by extreme parameter values.

\paragraph{Figure \ref{fig:heat_maps_lpg} Analysis:} Figure \ref{fig:heat_maps_lpg} presents the solvability heat maps for the LPG planner on domains with 15 objects, analyzed across a range of parameter settings. Like the FD, the LPG planner shows a decrease in solvability for configurations with high add/remove probabilities or negation values, indicating its sensitivity to complex domain setups. These findings emphasize LPG’s adaptability in certain configurations and reveal its struggles in randomized, diverse domains, further underscoring the necessity of diverse domain configurations for robust planner evaluation.

\subsection{Solvability Across Parameter Variations and Depth}

We evaluated the solvability of the generated domains across varying depths by conducting experiments with diverse domain configurations. Each configuration was defined by a unique combination of parameters, including probabilities for adding and removing precondition and effect predicates, negation probabilities, the number of objects, and predicate reversibility settings. Due to memory constraints on our system during the data generation process, the number of problems generated at each depth and the depths vary across configurations.

The tables below summarize the results, presenting the number of problems each planner (Fast Downward with FF and lmcut heuristics and LPG) successfully solves for different depths, parameter values, and domain characteristics. These results highlight the diversity and complexity of the generated domains, as even robust domain-independent planners face challenges in solving them, underscoring the effectiveness of our domain generator in creating complex domains.

\begin{table}[H]
\centering
\begin{tabular}{lccc}
    \toprule
    \textbf{Depth} & \textbf{FD(ff)} & \textbf{FD(lmcut)} & \textbf{LPG} \\
    \midrule
    1 & 5/5 & 5/5 & 2/5 \\
    2 & 5/5 & 5/5 & 0/5 \\
    3 & 5/5 & 5/5 & 0/5 \\
    4 & 1/1 & 1/1 & 0/1 \\

            \bottomrule
        \end{tabular}
        \caption{Solvability of Generated Domains for Depth 1 to 4, with probability values: prob\_add\_pre = 0.3, prob\_add\_eff = 0.3, prob\_rem\_pre = 0.7, prob\_rem\_eff = 0.7, negation = 0.3, predicate reversibility = True, number of objects = 5.0 }
        \end{table}


\begin{table}[H]
\centering
\begin{tabular}{lccc}
    \toprule
    \textbf{Depth} & \textbf{FD(ff)} & \textbf{FD(lmcut)} & \textbf{LPG} \\
    \midrule
    1 & 5/5 & 1/5 & 0/5 \\
    2 & 5/5 & 5/5 & 1/5 \\
    3 & 5/5 & 4/5 & 0/5 \\
    4 & 5/5 & 4/5 & 0/5 \\
    5 & 5/5 & 4/5 & 0/5 \\

            \bottomrule
        \end{tabular}
        \caption{Solvability of Generated Domains for Depth 1 to 5, with probability values: prob\_add\_pre = 0.3, prob\_add\_eff = 0.3, prob\_rem\_pre = 0.7, prob\_rem\_eff = 0.7, negation = 0.3, predicate reversibility = False, number of objects = 5.0 }
        \end{table}


\begin{table}[H]
\centering
\begin{tabular}{lccc}
    \toprule
    \textbf{Depth} & \textbf{FD(ff)} & \textbf{FD(lmcut)} & \textbf{LPG} \\
    \midrule
    1 & 5/5 & 1/5 & 3/5 \\
    2 & 5/5 & 4/5 & 1/5 \\
    3 & 3/5 & 3/5 & 0/5 \\
    4 & 3/3 & 3/3 & 0/3 \\

            \bottomrule
        \end{tabular}
        \caption{Solvability of Generated Domains for Depth 1 to 4, with probability values: prob\_add\_pre = 0.3, prob\_add\_eff = 0.3, prob\_rem\_pre = 0.7, prob\_rem\_eff = 0.7, negation = 0.5, predicate reversibility = True, number of objects = 5.0 }
        \end{table}


\begin{table}[H]
\centering
\begin{tabular}{lccc}
    \toprule
    \textbf{Depth} & \textbf{FD(ff)} & \textbf{FD(lmcut)} & \textbf{LPG} \\
    \midrule
    1 & 5/5 & 0/5 & 1/5 \\
    2 & 2/5 & 2/5 & 0/5 \\
    3 & 4/4 & 4/4 & 0/4 \\

            \bottomrule
        \end{tabular}
        \caption{Solvability of Generated Domains for Depth 1 to 3, with probability values: prob\_add\_pre = 0.3, prob\_add\_eff = 0.3, prob\_rem\_pre = 0.7, prob\_rem\_eff = 0.7, negation = 0.5, predicate reversibility = False, number of objects = 5.0 }
        \end{table}


\begin{table}[H]
\centering
\begin{tabular}{lccc}
    \toprule
    \textbf{Depth} & \textbf{FD(ff)} & \textbf{FD(lmcut)} & \textbf{LPG} \\
    \midrule
    1 & 3/5 & 1/5 & 5/5 \\
    2 & 2/5 & 4/5 & 2/5 \\
    3 & 4/5 & 4/5 & 0/5 \\
    4 & 4/5 & 4/5 & 1/5 \\

            \bottomrule
        \end{tabular}
        \caption{Solvability of Generated Domains for Depth 1 to 4, with probability values: prob\_add\_pre = 0.3, prob\_add\_eff = 0.3, prob\_rem\_pre = 0.7, prob\_rem\_eff = 0.7, negation = 0.7, predicate reversibility = True, number of objects = 5.0 }
        \end{table}


\begin{table}[H]
\centering
\begin{tabular}{lccc}
    \toprule
    \textbf{Depth} & \textbf{FD(ff)} & \textbf{FD(lmcut)} & \textbf{LPG} \\
    \midrule
    1 & 4/5 & 1/5 & 0/5 \\
    2 & 3/5 & 3/5 & 2/5 \\
    3 & 1/5 & 4/5 & 2/5 \\
    4 & 2/2 & 2/2 & 0/2 \\

            \bottomrule
        \end{tabular}
        \caption{Solvability of Generated Domains for Depth 1 to 4, with probability values: prob\_add\_pre = 0.3, prob\_add\_eff = 0.3, prob\_rem\_pre = 0.7, prob\_rem\_eff = 0.7, negation = 0.7, predicate reversibility = False, number of objects = 5.0 }
        \end{table}


\begin{table}[H]
\centering
\begin{tabular}{lccc}
    \toprule
    \textbf{Depth} & \textbf{FD(ff)} & \textbf{FD(lmcut)} & \textbf{LPG} \\
    \midrule
    1 & 1/5 & 2/5 & 1/5 \\
    2 & 1/5 & 3/5 & 1/5 \\
    3 & 5/5 & 5/5 & 0/5 \\
    4 & 5/5 & 4/5 & 0/5 \\
    5 & 2/2 & 2/2 & 0/2 \\

            \bottomrule
        \end{tabular}
        \caption{Solvability of Generated Domains for Depth 1 to 5, with probability values: prob\_add\_pre = 0.3, prob\_add\_eff = 0.5, prob\_rem\_pre = 0.7, prob\_rem\_eff = 0.5, negation = 0.3, predicate reversibility = True, number of objects = 5.0 }
        \end{table}


\begin{table}[H]
\centering
\begin{tabular}{lccc}
    \toprule
    \textbf{Depth} & \textbf{FD(ff)} & \textbf{FD(lmcut)} & \textbf{LPG} \\
    \midrule
    1 & 2/5 & 3/5 & 0/5 \\
    2 & 3/5 & 3/5 & 2/5 \\
    3 & 4/5 & 4/5 & 0/5 \\
    4 & 5/5 & 5/5 & 0/5 \\
    5 & 2/2 & 2/2 & 0/2 \\

            \bottomrule
        \end{tabular}
        \caption{Solvability of Generated Domains for Depth 1 to 5, with probability values: prob\_add\_pre = 0.3, prob\_add\_eff = 0.5, prob\_rem\_pre = 0.7, prob\_rem\_eff = 0.5, negation = 0.3, predicate reversibility = False, number of objects = 5.0 }
        \end{table}


\begin{table}[H]
\centering
\begin{tabular}{lccc}
    \toprule
    \textbf{Depth} & \textbf{FD(ff)} & \textbf{FD(lmcut)} & \textbf{LPG} \\
    \midrule
    1 & 2/5 & 3/5 & 5/5 \\
    2 & 4/5 & 4/5 & 0/5 \\
    3 & 5/5 & 5/5 & 0/5 \\
    4 & 0/1 & 0/1 & 0/1 \\

            \bottomrule
        \end{tabular}
        \caption{Solvability of Generated Domains for Depth 1 to 4, with probability values: prob\_add\_pre = 0.3, prob\_add\_eff = 0.5, prob\_rem\_pre = 0.7, prob\_rem\_eff = 0.5, negation = 0.5, predicate reversibility = True, number of objects = 5.0 }
        \end{table}


\begin{table}[H]
\centering
\begin{tabular}{lccc}
    \toprule
    \textbf{Depth} & \textbf{FD(ff)} & \textbf{FD(lmcut)} & \textbf{LPG} \\
    \midrule
    1 & 2/5 & 2/5 & 3/5 \\
    2 & 3/5 & 3/5 & 3/5 \\
    3 & 4/5 & 4/5 & 0/5 \\

            \bottomrule
        \end{tabular}
        \caption{Solvability of Generated Domains for Depth 1 to 3, with probability values: prob\_add\_pre = 0.3, prob\_add\_eff = 0.5, prob\_rem\_pre = 0.7, prob\_rem\_eff = 0.5, negation = 0.5, predicate reversibility = False, number of objects = 5.0 }
        \end{table}


\begin{table}[H]
\centering
\begin{tabular}{lccc}
    \toprule
    \textbf{Depth} & \textbf{FD(ff)} & \textbf{FD(lmcut)} & \textbf{LPG} \\
    \midrule
    1 & 2/5 & 1/5 & 3/5 \\
    2 & 2/5 & 2/5 & 1/5 \\
    3 & 2/3 & 2/3 & 0/3 \\

            \bottomrule
        \end{tabular}
        \caption{Solvability of Generated Domains for Depth 1 to 3, with probability values: prob\_add\_pre = 0.3, prob\_add\_eff = 0.5, prob\_rem\_pre = 0.7, prob\_rem\_eff = 0.5, negation = 0.7, predicate reversibility = True, number of objects = 5.0 }
        \end{table}


\begin{table}[H]
\centering
\begin{tabular}{lccc}
    \toprule
    \textbf{Depth} & \textbf{FD(ff)} & \textbf{FD(lmcut)} & \textbf{LPG} \\
    \midrule
    1 & 1/5 & 2/5 & 2/5 \\
    2 & 4/5 & 4/5 & 1/5 \\
    3 & 3/4 & 3/4 & 0/4 \\

            \bottomrule
        \end{tabular}
        \caption{Solvability of Generated Domains for Depth 1 to 3, with probability values: prob\_add\_pre = 0.3, prob\_add\_eff = 0.5, prob\_rem\_pre = 0.7, prob\_rem\_eff = 0.5, negation = 0.7, predicate reversibility = False, number of objects = 5.0 }
        \end{table}


\begin{table}[H]
\centering
\begin{tabular}{lccc}
    \toprule
    \textbf{Depth} & \textbf{FD(ff)} & \textbf{FD(lmcut)} & \textbf{LPG} \\
    \midrule
    1 & 3/5 & 1/5 & 2/5 \\
    2 & 3/5 & 3/5 & 1/5 \\
    3 & 4/4 & 4/4 & 0/4 \\

            \bottomrule
        \end{tabular}
        \caption{Solvability of Generated Domains for Depth 1 to 3, with probability values: prob\_add\_pre = 0.3, prob\_add\_eff = 0.7, prob\_rem\_pre = 0.7, prob\_rem\_eff = 0.3, negation = 0.3, predicate reversibility = True, number of objects = 5.0 }
        \end{table}


\begin{table}[H]
\centering
\begin{tabular}{lccc}
    \toprule
    \textbf{Depth} & \textbf{FD(ff)} & \textbf{FD(lmcut)} & \textbf{LPG} \\
    \midrule
    1 & 1/5 & 2/5 & 4/5 \\
    2 & 4/5 & 3/5 & 0/5 \\
    3 & 3/4 & 3/4 & 0/4 \\

            \bottomrule
        \end{tabular}
        \caption{Solvability of Generated Domains for Depth 1 to 3, with probability values: prob\_add\_pre = 0.3, prob\_add\_eff = 0.7, prob\_rem\_pre = 0.7, prob\_rem\_eff = 0.3, negation = 0.3, predicate reversibility = False, number of objects = 5.0 }
        \end{table}


\begin{table}[H]
\centering
\begin{tabular}{lccc}
    \toprule
    \textbf{Depth} & \textbf{FD(ff)} & \textbf{FD(lmcut)} & \textbf{LPG} \\
    \midrule
    1 & 4/5 & 3/5 & 2/5 \\
    2 & 3/5 & 4/5 & 0/5 \\

            \bottomrule
        \end{tabular}
        \caption{Solvability of Generated Domains for Depth 1 to 2, with probability values: prob\_add\_pre = 0.3, prob\_add\_eff = 0.7, prob\_rem\_pre = 0.7, prob\_rem\_eff = 0.3, negation = 0.5, predicate reversibility = True, number of objects = 5.0 }
        \end{table}


\begin{table}[H]
\centering
\begin{tabular}{lccc}
    \toprule
    \textbf{Depth} & \textbf{FD(ff)} & \textbf{FD(lmcut)} & \textbf{LPG} \\
    \midrule
    1 & 0/5 & 2/5 & 3/5 \\
    2 & 4/5 & 4/5 & 0/5 \\
    3 & 2/4 & 2/4 & 0/4 \\

            \bottomrule
        \end{tabular}
        \caption{Solvability of Generated Domains for Depth 1 to 3, with probability values: prob\_add\_pre = 0.3, prob\_add\_eff = 0.7, prob\_rem\_pre = 0.7, prob\_rem\_eff = 0.3, negation = 0.5, predicate reversibility = False, number of objects = 5.0 }
        \end{table}


\begin{table}[H]
\centering
\begin{tabular}{lccc}
    \toprule
    \textbf{Depth} & \textbf{FD(ff)} & \textbf{FD(lmcut)} & \textbf{LPG} \\
    \midrule
    1 & 0/5 & 0/5 & 1/5 \\
    2 & 3/5 & 3/5 & 0/5 \\
    3 & 4/5 & 4/5 & 0/5 \\

            \bottomrule
        \end{tabular}
        \caption{Solvability of Generated Domains for Depth 1 to 3, with probability values: prob\_add\_pre = 0.3, prob\_add\_eff = 0.7, prob\_rem\_pre = 0.7, prob\_rem\_eff = 0.3, negation = 0.7, predicate reversibility = True, number of objects = 5.0 }
        \end{table}


\begin{table}[H]
\centering
\begin{tabular}{lccc}
    \toprule
    \textbf{Depth} & \textbf{FD(ff)} & \textbf{FD(lmcut)} & \textbf{LPG} \\
    \midrule
    1 & 2/5 & 3/5 & 3/5 \\
    2 & 2/5 & 2/5 & 0/5 \\
    3 & 3/4 & 3/4 & 0/4 \\

            \bottomrule
        \end{tabular}
        \caption{Solvability of Generated Domains for Depth 1 to 3, with probability values: prob\_add\_pre = 0.3, prob\_add\_eff = 0.7, prob\_rem\_pre = 0.7, prob\_rem\_eff = 0.3, negation = 0.7, predicate reversibility = False, number of objects = 5.0 }
        \end{table}


\begin{table}[H]
\centering
\begin{tabular}{lccc}
    \toprule
    \textbf{Depth} & \textbf{FD(ff)} & \textbf{FD(lmcut)} & \textbf{LPG} \\
    \midrule
    1 & 3/5 & 2/5 & 0/5 \\
    2 & 5/5 & 4/5 & 1/5 \\
    3 & 4/5 & 4/5 & 0/5 \\

            \bottomrule
        \end{tabular}
        \caption{Solvability of Generated Domains for Depth 1 to 3, with probability values: prob\_add\_pre = 0.5, prob\_add\_eff = 0.3, prob\_rem\_pre = 0.5, prob\_rem\_eff = 0.7, negation = 0.3, predicate reversibility = True, number of objects = 5.0 }
        \end{table}


\begin{table}[H]
\centering
\begin{tabular}{lccc}
    \toprule
    \textbf{Depth} & \textbf{FD(ff)} & \textbf{FD(lmcut)} & \textbf{LPG} \\
    \midrule
    1 & 3/5 & 1/5 & 1/5 \\
    2 & 3/5 & 3/5 & 1/5 \\
    3 & 5/5 & 4/5 & 0/5 \\
    4 & 4/5 & 4/5 & 0/5 \\
    5 & 1/1 & 0/1 & 0/1 \\

            \bottomrule
        \end{tabular}
        \caption{Solvability of Generated Domains for Depth 1 to 5, with probability values: prob\_add\_pre = 0.5, prob\_add\_eff = 0.3, prob\_rem\_pre = 0.5, prob\_rem\_eff = 0.7, negation = 0.3, predicate reversibility = False, number of objects = 5.0 }
        \end{table}


\begin{table}[H]
\centering
\begin{tabular}{lccc}
    \toprule
    \textbf{Depth} & \textbf{FD(ff)} & \textbf{FD(lmcut)} & \textbf{LPG} \\
    \midrule
    1 & 1/5 & 2/5 & 3/5 \\
    2 & 5/5 & 4/5 & 0/5 \\
    3 & 5/5 & 5/5 & 0/5 \\
    4 & 5/5 & 5/5 & 0/5 \\
    5 & 2/2 & 2/2 & 0/2 \\

            \bottomrule
        \end{tabular}
        \caption{Solvability of Generated Domains for Depth 1 to 5, with probability values: prob\_add\_pre = 0.5, prob\_add\_eff = 0.3, prob\_rem\_pre = 0.5, prob\_rem\_eff = 0.7, negation = 0.5, predicate reversibility = True, number of objects = 5.0 }
        \end{table}


\begin{table}[H]
\centering
\begin{tabular}{lccc}
    \toprule
    \textbf{Depth} & \textbf{FD(ff)} & \textbf{FD(lmcut)} & \textbf{LPG} \\
    \midrule
    1 & 1/5 & 1/5 & 1/5 \\
    2 & 4/5 & 4/5 & 0/5 \\
    3 & 5/5 & 4/5 & 0/5 \\
    4 & 2/3 & 3/3 & 0/3 \\

            \bottomrule
        \end{tabular}
        \caption{Solvability of Generated Domains for Depth 1 to 4, with probability values: prob\_add\_pre = 0.5, prob\_add\_eff = 0.3, prob\_rem\_pre = 0.5, prob\_rem\_eff = 0.7, negation = 0.5, predicate reversibility = False, number of objects = 5.0 }
        \end{table}


\begin{table}[H]
\centering
\begin{tabular}{lccc}
    \toprule
    \textbf{Depth} & \textbf{FD(ff)} & \textbf{FD(lmcut)} & \textbf{LPG} \\
    \midrule
    1 & 2/5 & 1/5 & 1/5 \\
    2 & 3/5 & 4/5 & 1/5 \\
    3 & 3/4 & 3/4 & 0/4 \\

            \bottomrule
        \end{tabular}
        \caption{Solvability of Generated Domains for Depth 1 to 3, with probability values: prob\_add\_pre = 0.5, prob\_add\_eff = 0.3, prob\_rem\_pre = 0.5, prob\_rem\_eff = 0.7, negation = 0.7, predicate reversibility = True, number of objects = 5.0 }
        \end{table}


\begin{table}[H]
\centering
\begin{tabular}{lccc}
    \toprule
    \textbf{Depth} & \textbf{FD(ff)} & \textbf{FD(lmcut)} & \textbf{LPG} \\
    \midrule
    1 & 1/5 & 2/5 & 2/5 \\
    2 & 5/5 & 3/5 & 0/5 \\
    3 & 2/4 & 3/4 & 0/4 \\

            \bottomrule
        \end{tabular}
        \caption{Solvability of Generated Domains for Depth 1 to 3, with probability values: prob\_add\_pre = 0.5, prob\_add\_eff = 0.3, prob\_rem\_pre = 0.5, prob\_rem\_eff = 0.7, negation = 0.7, predicate reversibility = False, number of objects = 5.0 }
        \end{table}


\begin{table}[H]
\centering
\begin{tabular}{lccc}
    \toprule
    \textbf{Depth} & \textbf{FD(ff)} & \textbf{FD(lmcut)} & \textbf{LPG} \\
    \midrule
    1 & 2/5 & 1/5 & 1/5 \\
    2 & 5/5 & 4/5 & 0/5 \\
    3 & 5/5 & 5/5 & 0/5 \\

            \bottomrule
        \end{tabular}
        \caption{Solvability of Generated Domains for Depth 1 to 3, with probability values: prob\_add\_pre = 0.5, prob\_add\_eff = 0.5, prob\_rem\_pre = 0.5, prob\_rem\_eff = 0.5, negation = 0.3, predicate reversibility = True, number of objects = 5.0 }
        \end{table}


\begin{table}[H]
\centering
\begin{tabular}{lccc}
    \toprule
    \textbf{Depth} & \textbf{FD(ff)} & \textbf{FD(lmcut)} & \textbf{LPG} \\
    \midrule
    1 & 0/5 & 1/5 & 0/5 \\
    2 & 5/5 & 5/5 & 0/5 \\
    3 & 5/5 & 5/5 & 0/5 \\

            \bottomrule
        \end{tabular}
        \caption{Solvability of Generated Domains for Depth 1 to 3, with probability values: prob\_add\_pre = 0.5, prob\_add\_eff = 0.5, prob\_rem\_pre = 0.5, prob\_rem\_eff = 0.5, negation = 0.3, predicate reversibility = False, number of objects = 5.0 }
        \end{table}


\begin{table}[H]
\centering
\begin{tabular}{lccc}
    \toprule
    \textbf{Depth} & \textbf{FD(ff)} & \textbf{FD(lmcut)} & \textbf{LPG} \\
    \midrule
    1 & 3/5 & 2/5 & 1/5 \\
    2 & 5/5 & 3/5 & 0/5 \\
    3 & 5/5 & 5/5 & 1/5 \\
    4 & 2/3 & 3/3 & 0/3 \\

            \bottomrule
        \end{tabular}
        \caption{Solvability of Generated Domains for Depth 1 to 4, with probability values: prob\_add\_pre = 0.5, prob\_add\_eff = 0.5, prob\_rem\_pre = 0.5, prob\_rem\_eff = 0.5, negation = 0.5, predicate reversibility = True, number of objects = 5.0 }
        \end{table}


\begin{table}[H]
\centering
\begin{tabular}{lccc}
    \toprule
    \textbf{Depth} & \textbf{FD(ff)} & \textbf{FD(lmcut)} & \textbf{LPG} \\
    \midrule
    1 & 4/5 & 1/5 & 0/5 \\
    2 & 3/5 & 4/5 & 1/5 \\
    3 & 4/4 & 4/4 & 0/4 \\

            \bottomrule
        \end{tabular}
        \caption{Solvability of Generated Domains for Depth 1 to 3, with probability values: prob\_add\_pre = 0.5, prob\_add\_eff = 0.5, prob\_rem\_pre = 0.5, prob\_rem\_eff = 0.5, negation = 0.5, predicate reversibility = False, number of objects = 5.0 }
        \end{table}


\begin{table}[H]
\centering
\begin{tabular}{lccc}
    \toprule
    \textbf{Depth} & \textbf{FD(ff)} & \textbf{FD(lmcut)} & \textbf{LPG} \\
    \midrule
    1 & 1/5 & 1/5 & 4/5 \\
    2 & 4/5 & 4/5 & 0/5 \\
    3 & 4/4 & 4/4 & 0/4 \\

            \bottomrule
        \end{tabular}
        \caption{Solvability of Generated Domains for Depth 1 to 3, with probability values: prob\_add\_pre = 0.5, prob\_add\_eff = 0.5, prob\_rem\_pre = 0.5, prob\_rem\_eff = 0.5, negation = 0.7, predicate reversibility = True, number of objects = 5.0 }
        \end{table}


\begin{table}[H]
\centering
\begin{tabular}{lccc}
    \toprule
    \textbf{Depth} & \textbf{FD(ff)} & \textbf{FD(lmcut)} & \textbf{LPG} \\
    \midrule
    1 & 2/5 & 1/5 & 3/5 \\
    2 & 3/5 & 3/5 & 0/5 \\
    3 & 4/5 & 4/5 & 0/5 \\

            \bottomrule
        \end{tabular}
        \caption{Solvability of Generated Domains for Depth 1 to 3, with probability values: prob\_add\_pre = 0.5, prob\_add\_eff = 0.5, prob\_rem\_pre = 0.5, prob\_rem\_eff = 0.5, negation = 0.7, predicate reversibility = False, number of objects = 5.0 }
        \end{table}


\begin{table}[H]
\centering
\begin{tabular}{lccc}
    \toprule
    \textbf{Depth} & \textbf{FD(ff)} & \textbf{FD(lmcut)} & \textbf{LPG} \\
    \midrule
    1 & 3/5 & 2/5 & 1/5 \\
    2 & 3/5 & 3/5 & 0/5 \\
    3 & 5/5 & 4/5 & 0/5 \\
    4 & 3/3 & 3/3 & 0/3 \\

            \bottomrule
        \end{tabular}
        \caption{Solvability of Generated Domains for Depth 1 to 4, with probability values: prob\_add\_pre = 0.5, prob\_add\_eff = 0.7, prob\_rem\_pre = 0.5, prob\_rem\_eff = 0.3, negation = 0.3, predicate reversibility = True, number of objects = 5.0 }
        \end{table}


\begin{table}[H]
\centering
\begin{tabular}{lccc}
    \toprule
    \textbf{Depth} & \textbf{FD(ff)} & \textbf{FD(lmcut)} & \textbf{LPG} \\
    \midrule
    1 & 1/5 & 2/5 & 2/5 \\
    2 & 4/5 & 3/5 & 0/5 \\
    3 & 5/5 & 5/5 & 0/5 \\
    4 & 5/5 & 5/5 & 0/5 \\

            \bottomrule
        \end{tabular}
        \caption{Solvability of Generated Domains for Depth 1 to 4, with probability values: prob\_add\_pre = 0.5, prob\_add\_eff = 0.7, prob\_rem\_pre = 0.5, prob\_rem\_eff = 0.3, negation = 0.3, predicate reversibility = False, number of objects = 5.0 }
        \end{table}


\begin{table}[H]
\centering
\begin{tabular}{lccc}
    \toprule
    \textbf{Depth} & \textbf{FD(ff)} & \textbf{FD(lmcut)} & \textbf{LPG} \\
    \midrule
    1 & 1/5 & 1/5 & 4/5 \\
    2 & 4/5 & 4/5 & 0/5 \\
    3 & 4/5 & 5/5 & 0/5 \\

            \bottomrule
        \end{tabular}
        \caption{Solvability of Generated Domains for Depth 1 to 3, with probability values: prob\_add\_pre = 0.5, prob\_add\_eff = 0.7, prob\_rem\_pre = 0.5, prob\_rem\_eff = 0.3, negation = 0.5, predicate reversibility = True, number of objects = 5.0 }
        \end{table}


\begin{table}[H]
\centering
\begin{tabular}{lccc}
    \toprule
    \textbf{Depth} & \textbf{FD(ff)} & \textbf{FD(lmcut)} & \textbf{LPG} \\
    \midrule
    1 & 1/5 & 1/5 & 2/5 \\
    2 & 5/5 & 5/5 & 0/5 \\
    3 & 1/1 & 1/1 & 0/1 \\

            \bottomrule
        \end{tabular}
        \caption{Solvability of Generated Domains for Depth 1 to 3, with probability values: prob\_add\_pre = 0.5, prob\_add\_eff = 0.7, prob\_rem\_pre = 0.5, prob\_rem\_eff = 0.3, negation = 0.5, predicate reversibility = False, number of objects = 5.0 }
        \end{table}


\begin{table}[H]
\centering
\begin{tabular}{lccc}
    \toprule
    \textbf{Depth} & \textbf{FD(ff)} & \textbf{FD(lmcut)} & \textbf{LPG} \\
    \midrule
    1 & 2/5 & 0/5 & 4/5 \\
    2 & 3/5 & 3/5 & 0/5 \\

            \bottomrule
        \end{tabular}
        \caption{Solvability of Generated Domains for Depth 1 to 2, with probability values: prob\_add\_pre = 0.5, prob\_add\_eff = 0.7, prob\_rem\_pre = 0.5, prob\_rem\_eff = 0.3, negation = 0.7, predicate reversibility = True, number of objects = 5.0 }
        \end{table}


\begin{table}[H]
\centering
\begin{tabular}{lccc}
    \toprule
    \textbf{Depth} & \textbf{FD(ff)} & \textbf{FD(lmcut)} & \textbf{LPG} \\
    \midrule
    1 & 1/5 & 1/5 & 2/5 \\
    2 & 5/5 & 4/5 & 0/5 \\
    3 & 2/4 & 2/4 & 0/4 \\

            \bottomrule
        \end{tabular}
        \caption{Solvability of Generated Domains for Depth 1 to 3, with probability values: prob\_add\_pre = 0.5, prob\_add\_eff = 0.7, prob\_rem\_pre = 0.5, prob\_rem\_eff = 0.3, negation = 0.7, predicate reversibility = False, number of objects = 5.0 }
        \end{table}


\begin{table}[H]
\centering
\begin{tabular}{lccc}
    \toprule
    \textbf{Depth} & \textbf{FD(ff)} & \textbf{FD(lmcut)} & \textbf{LPG} \\
    \midrule
    1 & 0/5 & 1/5 & 0/5 \\
    2 & 4/5 & 3/5 & 0/5 \\
    3 & 1/1 & 1/1 & 0/1 \\

            \bottomrule
        \end{tabular}
        \caption{Solvability of Generated Domains for Depth 1 to 3, with probability values: prob\_add\_pre = 0.7, prob\_add\_eff = 0.3, prob\_rem\_pre = 0.3, prob\_rem\_eff = 0.7, negation = 0.3, predicate reversibility = True, number of objects = 5.0 }
        \end{table}


\begin{table}[H]
\centering
\begin{tabular}{lccc}
    \toprule
    \textbf{Depth} & \textbf{FD(ff)} & \textbf{FD(lmcut)} & \textbf{LPG} \\
    \midrule
    1 & 2/5 & 2/5 & 0/5 \\
    2 & 4/5 & 3/5 & 0/5 \\

            \bottomrule
        \end{tabular}
        \caption{Solvability of Generated Domains for Depth 1 to 2, with probability values: prob\_add\_pre = 0.7, prob\_add\_eff = 0.3, prob\_rem\_pre = 0.3, prob\_rem\_eff = 0.7, negation = 0.3, predicate reversibility = False, number of objects = 5.0 }
        \end{table}


\begin{table}[H]
\centering
\begin{tabular}{lccc}
    \toprule
    \textbf{Depth} & \textbf{FD(ff)} & \textbf{FD(lmcut)} & \textbf{LPG} \\
    \midrule
    1 & 2/5 & 2/5 & 1/5 \\
    2 & 3/5 & 4/5 & 0/5 \\
    3 & 4/4 & 2/4 & 0/4 \\

            \bottomrule
        \end{tabular}
        \caption{Solvability of Generated Domains for Depth 1 to 3, with probability values: prob\_add\_pre = 0.7, prob\_add\_eff = 0.3, prob\_rem\_pre = 0.3, prob\_rem\_eff = 0.7, negation = 0.5, predicate reversibility = True, number of objects = 5.0 }
        \end{table}


\begin{table}[H]
\centering
\begin{tabular}{lccc}
    \toprule
    \textbf{Depth} & \textbf{FD(ff)} & \textbf{FD(lmcut)} & \textbf{LPG} \\
    \midrule
    1 & 1/5 & 1/5 & 0/5 \\
    2 & 3/5 & 3/5 & 0/5 \\
    3 & 4/4 & 3/4 & 0/4 \\

            \bottomrule
        \end{tabular}
        \caption{Solvability of Generated Domains for Depth 1 to 3, with probability values: prob\_add\_pre = 0.7, prob\_add\_eff = 0.3, prob\_rem\_pre = 0.3, prob\_rem\_eff = 0.7, negation = 0.5, predicate reversibility = False, number of objects = 5.0 }
        \end{table}


\begin{table}[H]
\centering
\begin{tabular}{lccc}
    \toprule
    \textbf{Depth} & \textbf{FD(ff)} & \textbf{FD(lmcut)} & \textbf{LPG} \\
    \midrule
    1 & 2/5 & 4/5 & 0/5 \\
    2 & 5/5 & 4/5 & 0/5 \\
    3 & 4/5 & 5/5 & 0/5 \\
    4 & 2/2 & 2/2 & 0/2 \\

            \bottomrule
        \end{tabular}
        \caption{Solvability of Generated Domains for Depth 1 to 4, with probability values: prob\_add\_pre = 0.7, prob\_add\_eff = 0.3, prob\_rem\_pre = 0.3, prob\_rem\_eff = 0.7, negation = 0.7, predicate reversibility = True, number of objects = 5.0 }
        \end{table}


\begin{table}[H]
\centering
\begin{tabular}{lccc}
    \toprule
    \textbf{Depth} & \textbf{FD(ff)} & \textbf{FD(lmcut)} & \textbf{LPG} \\
    \midrule
    1 & 0/5 & 2/5 & 1/5 \\
    2 & 5/5 & 3/5 & 0/5 \\

            \bottomrule
        \end{tabular}
        \caption{Solvability of Generated Domains for Depth 1 to 2, with probability values: prob\_add\_pre = 0.7, prob\_add\_eff = 0.3, prob\_rem\_pre = 0.3, prob\_rem\_eff = 0.7, negation = 0.7, predicate reversibility = False, number of objects = 5.0 }
        \end{table}


\begin{table}[H]
\centering
\begin{tabular}{lccc}
    \toprule
    \textbf{Depth} & \textbf{FD(ff)} & \textbf{FD(lmcut)} & \textbf{LPG} \\
    \midrule
    1 & 1/5 & 1/5 & 0/5 \\
    2 & 5/5 & 3/5 & 0/5 \\
    3 & 5/5 & 4/5 & 0/5 \\

            \bottomrule
        \end{tabular}
        \caption{Solvability of Generated Domains for Depth 1 to 3, with probability values: prob\_add\_pre = 0.7, prob\_add\_eff = 0.5, prob\_rem\_pre = 0.3, prob\_rem\_eff = 0.5, negation = 0.3, predicate reversibility = True, number of objects = 5.0 }
        \end{table}


\begin{table}[H]
\centering
\begin{tabular}{lccc}
    \toprule
    \textbf{Depth} & \textbf{FD(ff)} & \textbf{FD(lmcut)} & \textbf{LPG} \\
    \midrule
    1 & 2/5 & 2/5 & 0/5 \\
    2 & 3/5 & 3/5 & 0/5 \\
    3 & 4/4 & 4/4 & 0/4 \\

            \bottomrule
        \end{tabular}
        \caption{Solvability of Generated Domains for Depth 1 to 3, with probability values: prob\_add\_pre = 0.7, prob\_add\_eff = 0.5, prob\_rem\_pre = 0.3, prob\_rem\_eff = 0.5, negation = 0.3, predicate reversibility = False, number of objects = 5.0 }
        \end{table}


\begin{table}[H]
\centering
\begin{tabular}{lccc}
    \toprule
    \textbf{Depth} & \textbf{FD(ff)} & \textbf{FD(lmcut)} & \textbf{LPG} \\
    \midrule
    1 & 2/5 & 1/5 & 2/5 \\
    2 & 2/5 & 3/5 & 0/5 \\
    3 & 5/5 & 5/5 & 0/5 \\
    4 & 5/5 & 5/5 & 0/5 \\

            \bottomrule
        \end{tabular}
        \caption{Solvability of Generated Domains for Depth 1 to 4, with probability values: prob\_add\_pre = 0.7, prob\_add\_eff = 0.5, prob\_rem\_pre = 0.3, prob\_rem\_eff = 0.5, negation = 0.5, predicate reversibility = True, number of objects = 5.0 }
        \end{table}


\begin{table}[H]
\centering
\begin{tabular}{lccc}
    \toprule
    \textbf{Depth} & \textbf{FD(ff)} & \textbf{FD(lmcut)} & \textbf{LPG} \\
    \midrule
    1 & 3/5 & 3/5 & 0/5 \\
    2 & 4/5 & 4/5 & 0/5 \\
    3 & 5/5 & 5/5 & 0/5 \\
    4 & 3/3 & 3/3 & 0/3 \\

            \bottomrule
        \end{tabular}
        \caption{Solvability of Generated Domains for Depth 1 to 4, with probability values: prob\_add\_pre = 0.7, prob\_add\_eff = 0.5, prob\_rem\_pre = 0.3, prob\_rem\_eff = 0.5, negation = 0.5, predicate reversibility = False, number of objects = 5.0 }
        \end{table}


\begin{table}[H]
\centering
\begin{tabular}{lccc}
    \toprule
    \textbf{Depth} & \textbf{FD(ff)} & \textbf{FD(lmcut)} & \textbf{LPG} \\
    \midrule
    1 & 0/5 & 3/5 & 2/5 \\
    2 & 2/5 & 3/5 & 0/5 \\

            \bottomrule
        \end{tabular}
        \caption{Solvability of Generated Domains for Depth 1 to 2, with probability values: prob\_add\_pre = 0.7, prob\_add\_eff = 0.5, prob\_rem\_pre = 0.3, prob\_rem\_eff = 0.5, negation = 0.7, predicate reversibility = True, number of objects = 5.0 }
        \end{table}


\begin{table}[H]
\centering
\begin{tabular}{lccc}
    \toprule
    \textbf{Depth} & \textbf{FD(ff)} & \textbf{FD(lmcut)} & \textbf{LPG} \\
    \midrule
    1 & 1/5 & 0/5 & 1/5 \\
    2 & 3/5 & 0/5 & 0/5 \\
    3 & 5/5 & 0/5 & 0/5 \\

            \bottomrule
        \end{tabular}
        \caption{Solvability of Generated Domains for Depth 1 to 3, with probability values: prob\_add\_pre = 0.7, prob\_add\_eff = 0.5, prob\_rem\_pre = 0.3, prob\_rem\_eff = 0.5, negation = 0.7, predicate reversibility = False, number of objects = 5.0 }
        \end{table}


\begin{table}[H]
\centering
\begin{tabular}{lccc}
    \toprule
    \textbf{Depth} & \textbf{FD(ff)} & \textbf{FD(lmcut)} & \textbf{LPG} \\
    \midrule
    1 & 3/5 & 0/5 & 0/5 \\
    2 & 4/5 & 0/5 & 0/5 \\

            \bottomrule
        \end{tabular}
        \caption{Solvability of Generated Domains for Depth 1 to 2, with probability values: prob\_add\_pre = 0.7, prob\_add\_eff = 0.7, prob\_rem\_pre = 0.3, prob\_rem\_eff = 0.3, negation = 0.3, predicate reversibility = True, number of objects = 5.0 }
        \end{table}


\begin{table}[H]
\centering
\begin{tabular}{lccc}
    \toprule
    \textbf{Depth} & \textbf{FD(ff)} & \textbf{FD(lmcut)} & \textbf{LPG} \\
    \midrule
    1 & 2/5 & 0/5 & 0/5 \\
    2 & 3/5 & 0/5 & 0/5 \\
    3 & 4/4 & 0/4 & 0/4 \\

            \bottomrule
        \end{tabular}
        \caption{Solvability of Generated Domains for Depth 1 to 3, with probability values: prob\_add\_pre = 0.7, prob\_add\_eff = 0.7, prob\_rem\_pre = 0.3, prob\_rem\_eff = 0.3, negation = 0.3, predicate reversibility = False, number of objects = 5.0 }
        \end{table}


\begin{table}[H]
\centering
\begin{tabular}{lccc}
    \toprule
    \textbf{Depth} & \textbf{FD(ff)} & \textbf{FD(lmcut)} & \textbf{LPG} \\
    \midrule
    1 & 3/5 & 5/5 & 2/5 \\
    2 & 4/5 & 3/5 & 0/5 \\
    3 & 3/4 & 0/4 & 0/4 \\

            \bottomrule
        \end{tabular}
        \caption{Solvability of Generated Domains for Depth 1 to 3, with probability values: prob\_add\_pre = 0.7, prob\_add\_eff = 0.7, prob\_rem\_pre = 0.3, prob\_rem\_eff = 0.3, negation = 0.5, predicate reversibility = True, number of objects = 5.0 }
        \end{table}


\begin{table}[H]
\centering
\begin{tabular}{lccc}
    \toprule
    \textbf{Depth} & \textbf{FD(ff)} & \textbf{FD(lmcut)} & \textbf{LPG} \\
    \midrule
    1 & 3/5 & 1/5 & 0/5 \\
    2 & 4/5 & 4/5 & 0/5 \\
    3 & 3/4 & 4/4 & 0/4 \\

            \bottomrule
        \end{tabular}
        \caption{Solvability of Generated Domains for Depth 1 to 3, with probability values: prob\_add\_pre = 0.7, prob\_add\_eff = 0.7, prob\_rem\_pre = 0.3, prob\_rem\_eff = 0.3, negation = 0.5, predicate reversibility = False, number of objects = 5.0 }
        \end{table}


\begin{table}[H]
\centering
\begin{tabular}{lccc}
    \toprule
    \textbf{Depth} & \textbf{FD(ff)} & \textbf{FD(lmcut)} & \textbf{LPG} \\
    \midrule
    1 & 3/5 & 1/5 & 0/5 \\
    2 & 5/5 & 4/5 & 0/5 \\
    3 & 4/4 & 4/4 & 0/4 \\

            \bottomrule
        \end{tabular}
        \caption{Solvability of Generated Domains for Depth 1 to 3, with probability values: prob\_add\_pre = 0.7, prob\_add\_eff = 0.7, prob\_rem\_pre = 0.3, prob\_rem\_eff = 0.3, negation = 0.7, predicate reversibility = True, number of objects = 5.0 }
        \end{table}


\begin{table}[H]
\centering
\begin{tabular}{lccc}
    \toprule
    \textbf{Depth} & \textbf{FD(ff)} & \textbf{FD(lmcut)} & \textbf{LPG} \\
    \midrule
    1 & 0/5 & 2/5 & 0/5 \\
    2 & 4/5 & 5/5 & 0/5 \\
    3 & 3/3 & 3/3 & 0/3 \\

            \bottomrule
        \end{tabular}
        \caption{Solvability of Generated Domains for Depth 1 to 3, with probability values: prob\_add\_pre = 0.7, prob\_add\_eff = 0.7, prob\_rem\_pre = 0.3, prob\_rem\_eff = 0.3, negation = 0.7, predicate reversibility = False, number of objects = 5.0 }
        \end{table}


\begin{table}[H]
\centering
\begin{tabular}{lccc}
    \toprule
    \textbf{Depth} & \textbf{FD(ff)} & \textbf{FD(lmcut)} & \textbf{LPG} \\
    \midrule
    1 & 3/5 & 5/5 & 1/5 \\
    2 & 5/5 & 5/5 & 0/5 \\
    3 & 3/3 & 3/3 & 0/3 \\

            \bottomrule
        \end{tabular}
        \caption{Solvability of Generated Domains for Depth 1 to 3, with probability values: prob\_add\_pre = 0.3, prob\_add\_eff = 0.3, prob\_rem\_pre = 0.7, prob\_rem\_eff = 0.7, negation = 0.3, predicate reversibility = True, number of objects = 10.0 }
        \end{table}


\begin{table}[H]
\centering
\begin{tabular}{lccc}
    \toprule
    \textbf{Depth} & \textbf{FD(ff)} & \textbf{FD(lmcut)} & \textbf{LPG} \\
    \midrule
    1 & 4/5 & 2/5 & 1/5 \\
    2 & 1/1 & 1/1 & 0/1 \\

            \bottomrule
        \end{tabular}
        \caption{Solvability of Generated Domains for Depth 1 to 2, with probability values: prob\_add\_pre = 0.3, prob\_add\_eff = 0.3, prob\_rem\_pre = 0.7, prob\_rem\_eff = 0.7, negation = 0.3, predicate reversibility = False, number of objects = 10.0 }
        \end{table}


\begin{table}[H]
\centering
\begin{tabular}{lccc}
    \toprule
    \textbf{Depth} & \textbf{FD(ff)} & \textbf{FD(lmcut)} & \textbf{LPG} \\
    \midrule
    1 & 2/5 & 1/5 & 3/5 \\
    2 & 2/5 & 2/5 & 1/5 \\

            \bottomrule
        \end{tabular}
        \caption{Solvability of Generated Domains for Depth 1 to 2, with probability values: prob\_add\_pre = 0.3, prob\_add\_eff = 0.3, prob\_rem\_pre = 0.7, prob\_rem\_eff = 0.7, negation = 0.5, predicate reversibility = True, number of objects = 10.0 }
        \end{table}


\begin{table}[H]
\centering
\begin{tabular}{lccc}
    \toprule
    \textbf{Depth} & \textbf{FD(ff)} & \textbf{FD(lmcut)} & \textbf{LPG} \\
    \midrule
    1 & 2/5 & 1/5 & 2/5 \\
    2 & 1/1 & 1/1 & 0/1 \\

            \bottomrule
        \end{tabular}
        \caption{Solvability of Generated Domains for Depth 1 to 2, with probability values: prob\_add\_pre = 0.3, prob\_add\_eff = 0.3, prob\_rem\_pre = 0.7, prob\_rem\_eff = 0.7, negation = 0.5, predicate reversibility = False, number of objects = 10.0 }
        \end{table}


\begin{table}[H]
\centering
\begin{tabular}{lccc}
    \toprule
    \textbf{Depth} & \textbf{FD(ff)} & \textbf{FD(lmcut)} & \textbf{LPG} \\
    \midrule
    1 & 4/5 & 3/5 & 2/5 \\
    2 & 4/5 & 4/5 & 1/5 \\

            \bottomrule
        \end{tabular}
        \caption{Solvability of Generated Domains for Depth 1 to 2, with probability values: prob\_add\_pre = 0.3, prob\_add\_eff = 0.3, prob\_rem\_pre = 0.7, prob\_rem\_eff = 0.7, negation = 0.7, predicate reversibility = True, number of objects = 10.0 }
        \end{table}


\begin{table}[H]
\centering
\begin{tabular}{lccc}
    \toprule
    \textbf{Depth} & \textbf{FD(ff)} & \textbf{FD(lmcut)} & \textbf{LPG} \\
    \midrule
    1 & 2/5 & 4/5 & 2/5 \\
    2 & 1/2 & 1/2 & 0/2 \\

            \bottomrule
        \end{tabular}
        \caption{Solvability of Generated Domains for Depth 1 to 2, with probability values: prob\_add\_pre = 0.3, prob\_add\_eff = 0.3, prob\_rem\_pre = 0.7, prob\_rem\_eff = 0.7, negation = 0.7, predicate reversibility = False, number of objects = 10.0 }
        \end{table}


\begin{table}[H]
\centering
\begin{tabular}{lccc}
    \toprule
    \textbf{Depth} & \textbf{FD(ff)} & \textbf{FD(lmcut)} & \textbf{LPG} \\
    \midrule
    1 & 2/5 & 3/5 & 1/5 \\
    2 & 4/5 & 4/5 & 0/5 \\

            \bottomrule
        \end{tabular}
        \caption{Solvability of Generated Domains for Depth 1 to 2, with probability values: prob\_add\_pre = 0.3, prob\_add\_eff = 0.5, prob\_rem\_pre = 0.7, prob\_rem\_eff = 0.5, negation = 0.3, predicate reversibility = True, number of objects = 10.0 }
        \end{table}


\begin{table}[H]
\centering
\begin{tabular}{lccc}
    \toprule
    \textbf{Depth} & \textbf{FD(ff)} & \textbf{FD(lmcut)} & \textbf{LPG} \\
    \midrule
    1 & 3/5 & 2/5 & 3/5 \\
    2 & 1/1 & 1/1 & 0/1 \\

            \bottomrule
        \end{tabular}
        \caption{Solvability of Generated Domains for Depth 1 to 2, with probability values: prob\_add\_pre = 0.3, prob\_add\_eff = 0.5, prob\_rem\_pre = 0.7, prob\_rem\_eff = 0.5, negation = 0.3, predicate reversibility = False, number of objects = 10.0 }
        \end{table}


\begin{table}[H]
\centering
\begin{tabular}{lccc}
    \toprule
    \textbf{Depth} & \textbf{FD(ff)} & \textbf{FD(lmcut)} & \textbf{LPG} \\
    \midrule
    1 & 3/5 & 2/5 & 1/5 \\
    2 & 3/5 & 3/5 & 0/5 \\

            \bottomrule
        \end{tabular}
        \caption{Solvability of Generated Domains for Depth 1 to 2, with probability values: prob\_add\_pre = 0.3, prob\_add\_eff = 0.5, prob\_rem\_pre = 0.7, prob\_rem\_eff = 0.5, negation = 0.5, predicate reversibility = True, number of objects = 10.0 }
        \end{table}


\begin{table}[H]
\centering
\begin{tabular}{lccc}
    \toprule
    \textbf{Depth} & \textbf{FD(ff)} & \textbf{FD(lmcut)} & \textbf{LPG} \\
    \midrule
    1 & 1/5 & 1/5 & 2/5 \\
    2 & 2/5 & 2/5 & 0/5 \\

            \bottomrule
        \end{tabular}
        \caption{Solvability of Generated Domains for Depth 1 to 2, with probability values: prob\_add\_pre = 0.3, prob\_add\_eff = 0.5, prob\_rem\_pre = 0.7, prob\_rem\_eff = 0.5, negation = 0.5, predicate reversibility = False, number of objects = 10.0 }
        \end{table}


\begin{table}[H]
\centering
\begin{tabular}{lccc}
    \toprule
    \textbf{Depth} & \textbf{FD(ff)} & \textbf{FD(lmcut)} & \textbf{LPG} \\
    \midrule
    1 & 1/5 & 2/5 & 1/5 \\
    2 & 0/1 & 0/1 & 0/1 \\

            \bottomrule
        \end{tabular}
        \caption{Solvability of Generated Domains for Depth 1 to 2, with probability values: prob\_add\_pre = 0.3, prob\_add\_eff = 0.5, prob\_rem\_pre = 0.7, prob\_rem\_eff = 0.5, negation = 0.7, predicate reversibility = True, number of objects = 10.0 }
        \end{table}


\begin{table}[H]
\centering
\begin{tabular}{lccc}
    \toprule
    \textbf{Depth} & \textbf{FD(ff)} & \textbf{FD(lmcut)} & \textbf{LPG} \\
    \midrule
    1 & 4/5 & 4/5 & 2/5 \\
    2 & 0/1 & 0/1 & 0/1 \\

            \bottomrule
        \end{tabular}
        \caption{Solvability of Generated Domains for Depth 1 to 2, with probability values: prob\_add\_pre = 0.3, prob\_add\_eff = 0.5, prob\_rem\_pre = 0.7, prob\_rem\_eff = 0.5, negation = 0.7, predicate reversibility = False, number of objects = 10.0 }
        \end{table}


\begin{table}[H]
\centering
\begin{tabular}{lccc}
    \toprule
    \textbf{Depth} & \textbf{FD(ff)} & \textbf{FD(lmcut)} & \textbf{LPG} \\
    \midrule
    1 & 2/5 & 1/5 & 2/5 \\
    2 & 0/1 & 0/1 & 0/1 \\

            \bottomrule
        \end{tabular}
        \caption{Solvability of Generated Domains for Depth 1 to 2, with probability values: prob\_add\_pre = 0.3, prob\_add\_eff = 0.7, prob\_rem\_pre = 0.7, prob\_rem\_eff = 0.3, negation = 0.3, predicate reversibility = True, number of objects = 10.0 }
        \end{table}


\begin{table}[H]
\centering
\begin{tabular}{lccc}
    \toprule
    \textbf{Depth} & \textbf{FD(ff)} & \textbf{FD(lmcut)} & \textbf{LPG} \\
    \midrule
    1 & 3/5 & 2/5 & 1/5 \\
    2 & 0/1 & 0/1 & 0/1 \\

            \bottomrule
        \end{tabular}
        \caption{Solvability of Generated Domains for Depth 1 to 2, with probability values: prob\_add\_pre = 0.3, prob\_add\_eff = 0.7, prob\_rem\_pre = 0.7, prob\_rem\_eff = 0.3, negation = 0.3, predicate reversibility = False, number of objects = 10.0 }
        \end{table}


\begin{table}[H]
\centering
\begin{tabular}{lccc}
    \toprule
    \textbf{Depth} & \textbf{FD(ff)} & \textbf{FD(lmcut)} & \textbf{LPG} \\
    \midrule
    1 & 1/5 & 2/5 & 1/5 \\
    2 & 0/1 & 0/1 & 0/1 \\

            \bottomrule
        \end{tabular}
        \caption{Solvability of Generated Domains for Depth 1 to 2, with probability values: prob\_add\_pre = 0.3, prob\_add\_eff = 0.7, prob\_rem\_pre = 0.7, prob\_rem\_eff = 0.3, negation = 0.5, predicate reversibility = True, number of objects = 10.0 }
        \end{table}


\begin{table}[H]
\centering
\begin{tabular}{lccc}
    \toprule
    \textbf{Depth} & \textbf{FD(ff)} & \textbf{FD(lmcut)} & \textbf{LPG} \\
    \midrule
    1 & 2/5 & 2/5 & 2/5 \\
    2 & 3/5 & 2/5 & 0/5 \\

            \bottomrule
        \end{tabular}
        \caption{Solvability of Generated Domains for Depth 1 to 2, with probability values: prob\_add\_pre = 0.3, prob\_add\_eff = 0.7, prob\_rem\_pre = 0.7, prob\_rem\_eff = 0.3, negation = 0.5, predicate reversibility = False, number of objects = 10.0 }
        \end{table}


\begin{table}[H]
\centering
\begin{tabular}{lccc}
    \toprule
    \textbf{Depth} & \textbf{FD(ff)} & \textbf{FD(lmcut)} & \textbf{LPG} \\
    \midrule
    1 & 3/5 & 1/5 & 1/5 \\
    2 & 0/1 & 0/1 & 0/1 \\

            \bottomrule
        \end{tabular}
        \caption{Solvability of Generated Domains for Depth 1 to 2, with probability values: prob\_add\_pre = 0.3, prob\_add\_eff = 0.7, prob\_rem\_pre = 0.7, prob\_rem\_eff = 0.3, negation = 0.7, predicate reversibility = True, number of objects = 10.0 }
        \end{table}


\begin{table}[H]
\centering
\begin{tabular}{lccc}
    \toprule
    \textbf{Depth} & \textbf{FD(ff)} & \textbf{FD(lmcut)} & \textbf{LPG} \\
    \midrule
    1 & 2/5 & 1/5 & 2/5 \\
    2 & 0/1 & 0/1 & 0/1 \\

            \bottomrule
        \end{tabular}
        \caption{Solvability of Generated Domains for Depth 1 to 2, with probability values: prob\_add\_pre = 0.3, prob\_add\_eff = 0.7, prob\_rem\_pre = 0.7, prob\_rem\_eff = 0.3, negation = 0.7, predicate reversibility = False, number of objects = 10.0 }
        \end{table}


\begin{table}[H]
\centering
\begin{tabular}{lccc}
    \toprule
    \textbf{Depth} & \textbf{FD(ff)} & \textbf{FD(lmcut)} & \textbf{LPG} \\
    \midrule
    1 & 1/5 & 2/5 & 1/5 \\
    2 & 4/5 & 4/5 & 1/5 \\

            \bottomrule
        \end{tabular}
        \caption{Solvability of Generated Domains for Depth 1 to 2, with probability values: prob\_add\_pre = 0.5, prob\_add\_eff = 0.3, prob\_rem\_pre = 0.5, prob\_rem\_eff = 0.7, negation = 0.3, predicate reversibility = True, number of objects = 10.0 }
        \end{table}


\begin{table}[H]
\centering
\begin{tabular}{lccc}
    \toprule
    \textbf{Depth} & \textbf{FD(ff)} & \textbf{FD(lmcut)} & \textbf{LPG} \\
    \midrule
    1 & 4/5 & 2/5 & 1/5 \\
    2 & 4/5 & 4/5 & 0/5 \\
    3 & 3/3 & 3/3 & 0/3 \\

            \bottomrule
        \end{tabular}
        \caption{Solvability of Generated Domains for Depth 1 to 3, with probability values: prob\_add\_pre = 0.5, prob\_add\_eff = 0.3, prob\_rem\_pre = 0.5, prob\_rem\_eff = 0.7, negation = 0.3, predicate reversibility = False, number of objects = 10.0 }
        \end{table}


\begin{table}[H]
\centering
\begin{tabular}{lccc}
    \toprule
    \textbf{Depth} & \textbf{FD(ff)} & \textbf{FD(lmcut)} & \textbf{LPG} \\
    \midrule
    1 & 3/5 & 3/5 & 2/5 \\
    2 & 1/1 & 0/1 & 0/1 \\

            \bottomrule
        \end{tabular}
        \caption{Solvability of Generated Domains for Depth 1 to 2, with probability values: prob\_add\_pre = 0.5, prob\_add\_eff = 0.3, prob\_rem\_pre = 0.5, prob\_rem\_eff = 0.7, negation = 0.5, predicate reversibility = True, number of objects = 10.0 }
        \end{table}


\begin{table}[H]
\centering
\begin{tabular}{lccc}
    \toprule
    \textbf{Depth} & \textbf{FD(ff)} & \textbf{FD(lmcut)} & \textbf{LPG} \\
    \midrule
    1 & 5/5 & 3/5 & 2/5 \\
    2 & 4/5 & 3/5 & 0/5 \\

            \bottomrule
        \end{tabular}
        \caption{Solvability of Generated Domains for Depth 1 to 2, with probability values: prob\_add\_pre = 0.5, prob\_add\_eff = 0.3, prob\_rem\_pre = 0.5, prob\_rem\_eff = 0.7, negation = 0.5, predicate reversibility = False, number of objects = 10.0 }
        \end{table}


\begin{table}[H]
\centering
\begin{tabular}{lccc}
    \toprule
    \textbf{Depth} & \textbf{FD(ff)} & \textbf{FD(lmcut)} & \textbf{LPG} \\
    \midrule
    1 & 3/5 & 1/5 & 2/5 \\
    2 & 1/1 & 1/1 & 1/1 \\

            \bottomrule
        \end{tabular}
        \caption{Solvability of Generated Domains for Depth 1 to 2, with probability values: prob\_add\_pre = 0.5, prob\_add\_eff = 0.3, prob\_rem\_pre = 0.5, prob\_rem\_eff = 0.7, negation = 0.7, predicate reversibility = True, number of objects = 10.0 }
        \end{table}


\begin{table}[H]
\centering
\begin{tabular}{lccc}
    \toprule
    \textbf{Depth} & \textbf{FD(ff)} & \textbf{FD(lmcut)} & \textbf{LPG} \\
    \midrule
    1 & 3/5 & 3/5 & 2/5 \\
    2 & 1/1 & 1/1 & 0/1 \\

            \bottomrule
        \end{tabular}
        \caption{Solvability of Generated Domains for Depth 1 to 2, with probability values: prob\_add\_pre = 0.5, prob\_add\_eff = 0.3, prob\_rem\_pre = 0.5, prob\_rem\_eff = 0.7, negation = 0.7, predicate reversibility = False, number of objects = 10.0 }
        \end{table}


\begin{table}[H]
\centering
\begin{tabular}{lccc}
    \toprule
    \textbf{Depth} & \textbf{FD(ff)} & \textbf{FD(lmcut)} & \textbf{LPG} \\
    \midrule
    1 & 3/5 & 3/5 & 1/5 \\
    2 & 3/5 & 2/5 & 0/5 \\

            \bottomrule
        \end{tabular}
        \caption{Solvability of Generated Domains for Depth 1 to 2, with probability values: prob\_add\_pre = 0.5, prob\_add\_eff = 0.5, prob\_rem\_pre = 0.5, prob\_rem\_eff = 0.5, negation = 0.3, predicate reversibility = True, number of objects = 10.0 }
        \end{table}


\begin{table}[H]
\centering
\begin{tabular}{lccc}
    \toprule
    \textbf{Depth} & \textbf{FD(ff)} & \textbf{FD(lmcut)} & \textbf{LPG} \\
    \midrule
    1 & 3/5 & 3/5 & 0/5 \\
    2 & 1/1 & 1/1 & 0/1 \\

            \bottomrule
        \end{tabular}
        \caption{Solvability of Generated Domains for Depth 1 to 2, with probability values: prob\_add\_pre = 0.5, prob\_add\_eff = 0.5, prob\_rem\_pre = 0.5, prob\_rem\_eff = 0.5, negation = 0.3, predicate reversibility = False, number of objects = 10.0 }
        \end{table}


\begin{table}[H]
\centering
\begin{tabular}{lccc}
    \toprule
    \textbf{Depth} & \textbf{FD(ff)} & \textbf{FD(lmcut)} & \textbf{LPG} \\
    \midrule
    1 & 3/5 & 3/5 & 3/5 \\
    2 & 1/1 & 1/1 & 0/1 \\

            \bottomrule
        \end{tabular}
        \caption{Solvability of Generated Domains for Depth 1 to 2, with probability values: prob\_add\_pre = 0.5, prob\_add\_eff = 0.5, prob\_rem\_pre = 0.5, prob\_rem\_eff = 0.5, negation = 0.5, predicate reversibility = True, number of objects = 10.0 }
        \end{table}


\begin{table}[H]
\centering
\begin{tabular}{lccc}
    \toprule
    \textbf{Depth} & \textbf{FD(ff)} & \textbf{FD(lmcut)} & \textbf{LPG} \\
    \midrule
    1 & 2/5 & 3/5 & 2/5 \\
    2 & 2/2 & 2/2 & 0/2 \\

            \bottomrule
        \end{tabular}
        \caption{Solvability of Generated Domains for Depth 1 to 2, with probability values: prob\_add\_pre = 0.5, prob\_add\_eff = 0.5, prob\_rem\_pre = 0.5, prob\_rem\_eff = 0.5, negation = 0.5, predicate reversibility = False, number of objects = 10.0 }
        \end{table}


\begin{table}[H]
\centering
\begin{tabular}{lccc}
    \toprule
    \textbf{Depth} & \textbf{FD(ff)} & \textbf{FD(lmcut)} & \textbf{LPG} \\
    \midrule
    1 & 3/5 & 4/5 & 0/5 \\
    2 & 1/1 & 1/1 & 0/1 \\

            \bottomrule
        \end{tabular}
        \caption{Solvability of Generated Domains for Depth 1 to 2, with probability values: prob\_add\_pre = 0.5, prob\_add\_eff = 0.5, prob\_rem\_pre = 0.5, prob\_rem\_eff = 0.5, negation = 0.7, predicate reversibility = True, number of objects = 10.0 }
        \end{table}


\begin{table}[H]
\centering
\begin{tabular}{lccc}
    \toprule
    \textbf{Depth} & \textbf{FD(ff)} & \textbf{FD(lmcut)} & \textbf{LPG} \\
    \midrule
    1 & 2/5 & 3/5 & 0/5 \\
    2 & 1/2 & 1/2 & 0/2 \\

            \bottomrule
        \end{tabular}
        \caption{Solvability of Generated Domains for Depth 1 to 2, with probability values: prob\_add\_pre = 0.5, prob\_add\_eff = 0.5, prob\_rem\_pre = 0.5, prob\_rem\_eff = 0.5, negation = 0.7, predicate reversibility = False, number of objects = 10.0 }
        \end{table}


\begin{table}[H]
\centering
\begin{tabular}{lccc}
    \toprule
    \textbf{Depth} & \textbf{FD(ff)} & \textbf{FD(lmcut)} & \textbf{LPG} \\
    \midrule
    1 & 1/5 & 0/5 & 2/5 \\
    2 & 1/1 & 1/1 & 0/1 \\

            \bottomrule
        \end{tabular}
        \caption{Solvability of Generated Domains for Depth 1 to 2, with probability values: prob\_add\_pre = 0.5, prob\_add\_eff = 0.7, prob\_rem\_pre = 0.5, prob\_rem\_eff = 0.3, negation = 0.3, predicate reversibility = True, number of objects = 10.0 }
        \end{table}


\begin{table}[H]
\centering
\begin{tabular}{lccc}
    \toprule
    \textbf{Depth} & \textbf{FD(ff)} & \textbf{FD(lmcut)} & \textbf{LPG} \\
    \midrule
    1 & 3/5 & 2/5 & 0/5 \\
    2 & 1/1 & 1/1 & 0/1 \\

            \bottomrule
        \end{tabular}
        \caption{Solvability of Generated Domains for Depth 1 to 2, with probability values: prob\_add\_pre = 0.5, prob\_add\_eff = 0.7, prob\_rem\_pre = 0.5, prob\_rem\_eff = 0.3, negation = 0.3, predicate reversibility = False, number of objects = 10.0 }
        \end{table}


\begin{table}[H]
\centering
\begin{tabular}{lccc}
    \toprule
    \textbf{Depth} & \textbf{FD(ff)} & \textbf{FD(lmcut)} & \textbf{LPG} \\
    \midrule
    1 & 3/5 & 3/5 & 0/5 \\
    2 & 1/1 & 1/1 & 0/1 \\

            \bottomrule
        \end{tabular}
        \caption{Solvability of Generated Domains for Depth 1 to 2, with probability values: prob\_add\_pre = 0.5, prob\_add\_eff = 0.7, prob\_rem\_pre = 0.5, prob\_rem\_eff = 0.3, negation = 0.5, predicate reversibility = True, number of objects = 10.0 }
        \end{table}


\begin{table}[H]
\centering
\begin{tabular}{lccc}
    \toprule
    \textbf{Depth} & \textbf{FD(ff)} & \textbf{FD(lmcut)} & \textbf{LPG} \\
    \midrule
    1 & 2/5 & 1/5 & 1/5 \\
    2 & 1/1 & 1/1 & 0/1 \\

            \bottomrule
        \end{tabular}
        \caption{Solvability of Generated Domains for Depth 1 to 2, with probability values: prob\_add\_pre = 0.5, prob\_add\_eff = 0.7, prob\_rem\_pre = 0.5, prob\_rem\_eff = 0.3, negation = 0.5, predicate reversibility = False, number of objects = 10.0 }
        \end{table}


\begin{table}[H]
\centering
\begin{tabular}{lccc}
    \toprule
    \textbf{Depth} & \textbf{FD(ff)} & \textbf{FD(lmcut)} & \textbf{LPG} \\
    \midrule
    1 & 4/5 & 3/5 & 0/5 \\
    2 & 0/1 & 0/1 & 0/1 \\

            \bottomrule
        \end{tabular}
        \caption{Solvability of Generated Domains for Depth 1 to 2, with probability values: prob\_add\_pre = 0.5, prob\_add\_eff = 0.7, prob\_rem\_pre = 0.5, prob\_rem\_eff = 0.3, negation = 0.7, predicate reversibility = True, number of objects = 10.0 }
        \end{table}


\begin{table}[H]
\centering
\begin{tabular}{lccc}
    \toprule
    \textbf{Depth} & \textbf{FD(ff)} & \textbf{FD(lmcut)} & \textbf{LPG} \\
    \midrule
    1 & 0/5 & 0/5 & 2/5 \\
    2 & 1/1 & 1/1 & 0/1 \\

            \bottomrule
        \end{tabular}
        \caption{Solvability of Generated Domains for Depth 1 to 2, with probability values: prob\_add\_pre = 0.5, prob\_add\_eff = 0.7, prob\_rem\_pre = 0.5, prob\_rem\_eff = 0.3, negation = 0.7, predicate reversibility = False, number of objects = 10.0 }
        \end{table}


\begin{table}[H]
\centering
\begin{tabular}{lccc}
    \toprule
    \textbf{Depth} & \textbf{FD(ff)} & \textbf{FD(lmcut)} & \textbf{LPG} \\
    \midrule
    1 & 3/5 & 1/5 & 0/5 \\
    2 & 1/1 & 1/1 & 0/1 \\

            \bottomrule
        \end{tabular}
        \caption{Solvability of Generated Domains for Depth 1 to 2, with probability values: prob\_add\_pre = 0.7, prob\_add\_eff = 0.3, prob\_rem\_pre = 0.3, prob\_rem\_eff = 0.7, negation = 0.3, predicate reversibility = True, number of objects = 10.0 }
        \end{table}


\begin{table}[H]
\centering
\begin{tabular}{lccc}
    \toprule
    \textbf{Depth} & \textbf{FD(ff)} & \textbf{FD(lmcut)} & \textbf{LPG} \\
    \midrule
    1 & 5/5 & 0/5 & 0/5 \\
    2 & 1/1 & 0/1 & 0/1 \\

            \bottomrule
        \end{tabular}
        \caption{Solvability of Generated Domains for Depth 1 to 2, with probability values: prob\_add\_pre = 0.7, prob\_add\_eff = 0.3, prob\_rem\_pre = 0.3, prob\_rem\_eff = 0.7, negation = 0.3, predicate reversibility = False, number of objects = 10.0 }
        \end{table}


\begin{table}[H]
\centering
\begin{tabular}{lccc}
    \toprule
    \textbf{Depth} & \textbf{FD(ff)} & \textbf{FD(lmcut)} & \textbf{LPG} \\
    \midrule
    1 & 2/5 & 2/5 & 2/5 \\
    2 & 2/2 & 2/2 & 0/2 \\

            \bottomrule
        \end{tabular}%
    \caption{Solvability of Generated Domains for Depth 1 , with probability values: prob\_add\_pre = 0.3, prob\_add\_eff = 0.3, prob\_rem\_pre = 0.7, prob\_rem\_eff = 0.7, negation = 0.3, predicate reversibility = True, number of objects = 15.0 }
    \end{table}


\begin{table}[H]
\centering
\begin{tabular}{lccc}
    \toprule
    \textbf{Depth} & \textbf{FD(ff)} & \textbf{FD(lmcut)} & \textbf{LPG} \\
    \midrule
    1 & 3/5 & 1/5 & 1/5 \\

        \bottomrule
    \end{tabular}
    \caption{Solvability of Generated Domains for Depth 1 , with probability values: prob\_add\_pre = 0.3, prob\_add\_eff = 0.3, prob\_rem\_pre = 0.7, prob\_rem\_eff = 0.7, negation = 0.3, predicate reversibility = False, number of objects = 15.0 }
    \end{table}


\begin{table}[H]
\centering
%
    \caption{Solvability of Generated Domains for Depth 1 , with probability values: prob\_add\_pre = 0.3, prob\_add\_eff = 0.3, prob\_rem\_pre = 0.7, prob\_rem\_eff = 0.7, negation = 0.5, predicate reversibility = False, number of objects = 15.0 }
    \end{table}


\begin{table}[H]
\centering
\begin{tabular}{lccc}
    \toprule
    \textbf{Depth} & \textbf{FD(ff)} & \textbf{FD(lmcut)} & \textbf{LPG} \\
    \midrule
    1 & 1/5 & 4/5 & 1/5 \\

        \bottomrule
    \end{tabular}
    \caption{Solvability of Generated Domains for Depth 1 , with probability values: prob\_add\_pre = 0.3, prob\_add\_eff = 0.3, prob\_rem\_pre = 0.7, prob\_rem\_eff = 0.7, negation = 0.7, predicate reversibility = True, number of objects = 15.0 }
    \end{table}


\begin{table}[H]
\centering
%
    \caption{Solvability of Generated Domains for Depth 1 , with probability values: prob\_add\_pre = 0.3, prob\_add\_eff = 0.3, prob\_rem\_pre = 0.7, prob\_rem\_eff = 0.7, negation = 0.7, predicate reversibility = False, number of objects = 15.0 }
    \end{table}


\begin{table}[H]
\centering
%
    \caption{Solvability of Generated Domains for Depth 1 , with probability values: prob\_add\_pre = 0.3, prob\_add\_eff = 0.5, prob\_rem\_pre = 0.7, prob\_rem\_eff = 0.5, negation = 0.3, predicate reversibility = True, number of objects = 15.0 }
    \end{table}


\begin{table}[H]
\centering
%
    \caption{Solvability of Generated Domains for Depth 1 , with probability values: prob\_add\_pre = 0.3, prob\_add\_eff = 0.3, prob\_rem\_pre = 0.7, prob\_rem\_eff = 0.7, negation = 0.3, predicate reversibility = True, number of objects = None }
    \end{table}


\begin{table}[H]
\centering
%
    \caption{Solvability of Generated Domains for Depth 1 , with probability values: prob\_add\_pre = 0.3, prob\_add\_eff = 0.3, prob\_rem\_pre = 0.7, prob\_rem\_eff = 0.7, negation = 0.3, predicate reversibility = False, number of objects = None }
    \end{table}


\begin{table}[H]
\centering
%
    \caption{Solvability of Generated Domains for Depth 1 , with probability values: prob\_add\_pre = 0.3, prob\_add\_eff = 0.3, prob\_rem\_pre = 0.7, prob\_rem\_eff = 0.7, negation = 0.5, predicate reversibility = True, number of objects = None }
    \end{table}


\begin{table}[H]
\centering
%
    \caption{Solvability of Generated Domains for Depth 1 , with probability values: prob\_add\_pre = 0.3, prob\_add\_eff = 0.3, prob\_rem\_pre = 0.7, prob\_rem\_eff = 0.7, negation = 0.5, predicate reversibility = False, number of objects = None }
    \end{table}


\begin{table}[H]
\centering
%
    \caption{Solvability of Generated Domains for Depth 1 , with probability values: prob\_add\_pre = 0.3, prob\_add\_eff = 0.3, prob\_rem\_pre = 0.7, prob\_rem\_eff = 0.7, negation = 0.7, predicate reversibility = True, number of objects = None }
    \end{table}


\begin{table}[H]
\centering
%
    \caption{Solvability of Generated Domains for Depth 1 , with probability values: prob\_add\_pre = 0.3, prob\_add\_eff = 0.3, prob\_rem\_pre = 0.7, prob\_rem\_eff = 0.7, negation = 0.7, predicate reversibility = False, number of objects = None }
    \end{table}


\begin{table}[H]
\centering
%
    \caption{Solvability of Generated Domains for Depth 1 , with probability values: prob\_add\_pre = 0.7, prob\_add\_eff = 0.7, prob\_rem\_pre = 0.3, prob\_rem\_eff = 0.3, negation = 0.7, predicate reversibility = False, number of objects = None }
    \end{table}

\end{document}